\documentclass[10pt,twocolumn,letterpaper]{article}
\usepackage{cvpr}              

\usepackage{dsfont}
\usepackage{graphicx}
\usepackage{amsmath}
\usepackage{amssymb}
\usepackage{booktabs}
\usepackage{lipsum}
\usepackage{enumitem} 
\usepackage{tikz}
\usepackage{pgfplots}
\usepackage{multirow}
\usepackage{makecell}
\usepackage{bm}

\usepackage[pagebackref,breaklinks,colorlinks]{hyperref}

\usepackage[capitalize]{cleveref}
\crefname{section}{Sec.}{Secs.}
\Crefname{section}{Section}{Sections}
\Crefname{table}{Table}{Tables}
\crefname{table}{Tab.}{Tabs.}

\def\cvprPaperID{0000} 
\def\confName{CVPR}
\def\confYear{2024}

\begin{document}

\title{InstructDiffusion: A Generalist Modeling Interface for Vision Tasks}

\author{Zigang Geng$\footnotemark[1]$, Binxin Yang$\footnotemark[1]$, Tiankai Hang$\footnotemark[1]$, Chen Li$\footnotemark[1]$, 
Shuyang Gu$\footnotemark[2]$, \\
Ting Zhang, Jianmin Bao, Zheng Zhang, Han Hu, Dong Chen, Baining Guo\\
Microsoft Research Asia\\
{\tt\small \url{https://gengzigang.github.io/instructdiffusion.github.io/}}}


\twocolumn[{%
\renewcommand\twocolumn[1][]{#1}%
\maketitle

\begin{center}
    \vspace{-1.0em}
    \centering
    \captionsetup{type=figure}
    \includegraphics[width=1\textwidth, trim={0 0 7 0},clip]{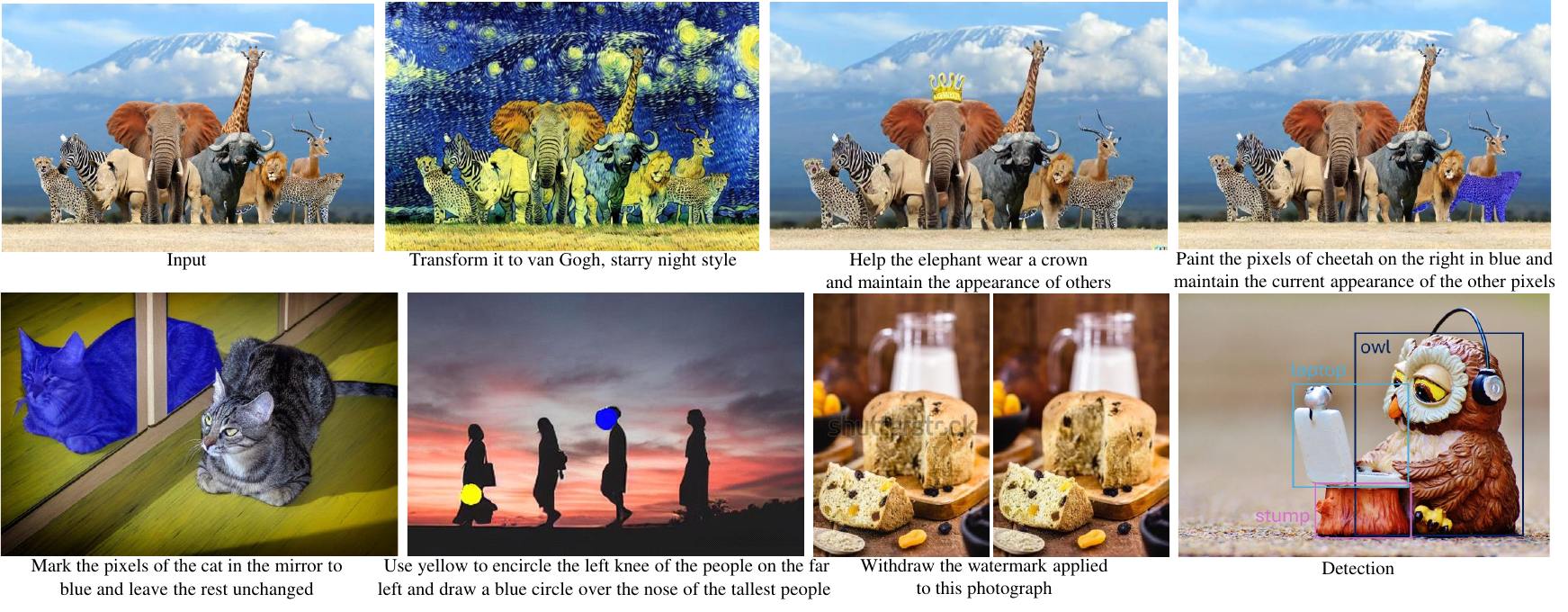}
    \captionof{figure}{We introduce InstructDiffusion, a generalist modeling interface for vision tasks. Given input image and human instruction, our unified model effectively accomplishes tasks such as image editing, segmentation, keypoint estimation, detection, and low-level vision. }
    \label{fig:teaser-candidate}
\end{center}%

}]

\renewcommand{\thefootnote}{\fnsymbol{footnote}}  
\footnotetext[1]{Equal contribution.}
\footnotetext[2]{Corresponding Author.}
\renewcommand{\thefootnote}{\arabic{footnote}}  

\begin{abstract}

We present InstructDiffusion, a unifying and generic framework for aligning computer vision tasks with human instructions. Unlike existing approaches that integrate prior knowledge and pre-define the output space (\eg, categories and coordinates) for each vision task, we cast diverse vision tasks into a human-intuitive image-manipulating process whose output space is a flexible and interactive pixel space. Concretely, the model is built upon the diffusion process and is trained to predict pixels according to user instructions, such as encircling the man's left shoulder in red or applying a blue mask to the left car. InstructDiffusion could handle a variety of vision tasks, including understanding tasks (such as segmentation and keypoint detection) and generative tasks (such as editing and enhancement). It even exhibits the ability to handle unseen tasks and outperforms prior methods on novel datasets. This represents a significant step towards a generalist modeling interface for vision tasks, advancing artificial general intelligence in the field of computer vision.

\end{abstract}

\section{Introduction}
\label{sec:intro}

In recent years, the field of artificial intelligence has witnessed remarkable advancements, particularly in natural language processing (NLP)~\cite{devlin2018bert,brown2020language,ouyang2022training,openai2023gpt4}. The Generative Pre-trained Transformer (GPT) has successfully unified multiple NLP tasks by providing a single, coherent framework for diverse applications. Building on this success, our research aims to achieve a similar unification in the realm of computer vision, \ie~\cite{chen2021pix2seq,chen2022unified}, developing a unifying framework capable of handling multiple vision tasks simultaneously. However, compared with NLP tasks, unifying computer vision tasks is more challenging due to the diversity of various tasks.

\noindent\textbf{Diversity of Tasks and Outputs}: Computer vision tasks encompass a wide range of applications, such as object recognition, segmentation, image generation, and keypoint detection, among others. Each of these tasks has a different output format, including coordinates, binary masks, images, and categories. This diversity makes it difficult to find a uniform representation for all tasks. In contrast, NLP tasks often have text-based outputs that can be more easily represented in a standard format.

\noindent\textbf{Different Methodologies and Techniques}: Computer vision tasks often require distinct methodologies and techniques depending on the specific problem being addressed. For example, image generation tasks are commonly dominated by Generative Adversarial Networks (GANs)~\cite{goodfellow2020generative,karras2019style,karras2020analyzing} and Denoising Diffusion Models (DDPM)~\cite{ho2020denoising,song2020score,gu2022vector}, which are rarely used for image understanding tasks such as object recognition or image classification. Additionally, the output dimensionality of generative models is relatively higher, adding to the challenge of unifying these tasks. In contrast, NLP tasks tend to rely on a more consistent set of techniques, such as Transformer-based models~\cite{vaswani2017attention}, which can be applied across various NLP applications.

\noindent\textbf{Continuous Input and Output}: Both the input and output of computer vision tasks are usually continuous, like coordinates or images. This continuous nature makes it challenging to develop a unified approach that can accurately handle such data. If discretizing the continuous data using techniques like Vector Quantized-Variational AutoEncoders (VQ-VAE)~\cite{van2017neural,razavi2019generating}, there will be quantization errors, leading to inaccuracies in the results. This issue is less prominent in NLP tasks, where the input and output data can be more easily discretized~\cite{vaswani2017attention,devlin2018bert,brown2020language} into text tokens.



In this paper, we take advantage of the DDPM and propose a novel approach to address these challenges by treating all computer vision tasks as image generation, specifically instructional image editing tasks. We instruct image editing tasks using a more natural and intuitive way that closely aligns with how humans process images. For instance, the instruction for a segmentation task could involve turning the pixels of an object in the image into a specific color, while the remaining pixels remain unchanged. The keypoint detection task can be described as placing an opaque colored circle at a specific position in the image. The instruction for a classification task could change the object to different colors according to its category. Compared with some methods that have attempted to formulate vision tasks as inpainting problems~\cite{wang2023images,bar2022visual}, our approach ensures accurate reflection of human intentions which simplifies the process of handling multiple vision tasks. At the same time, since the input and output of DDPM are continuous~\cite{ho2020denoising,song2020score}, discretization is unnecessary, which solves the problem of quantization error.


We mainly focus on three types of output formats: 3-channel RGB images, binary masks, and keypoints. These three outputs are sufficient to cover most vision tasks, such as semantic segmentation, referring segmentation, keypoint detection, image manipulation, and so on. Since the output of the denoising diffusion model is a 3-channel image, we propose a unified representation that encodes masks and keypoints into 3-channel images to handle various image understanding tasks. Then we use a post-processing module to extract the commonly used output format for evaluation.

During the training phase, we use a diverse set of tasks to train a single model uniformly. We also collect a new dataset for image editing. The experimental results demonstrate that our approach achieves good performance in each task. Furthermore, we observed that, compared to training individual models for each task, joint training of multiple tasks can enhance the generalization ability.

Remarkably, our model also exhibits the ability of AGI to a certain extent, as it can handle tasks not seen during the training phase, such as image detection and classification. Moreover, it performs better than previous methods on datasets that were not seen during training. This study thus presents a significant step towards the development of a generalist modeling interface for vision tasks, paving the way for future research in the quest for AGI in computer vision.

\section{Related Work}
\label{sec:related_work}

\noindent
Building a general-purpose model that is capable of solving any arbitrary task has been a longstanding desire for artificial intelligence research. There exists a substantial number of related works in the literature, aiming to unify a broad spectrum of tasks. We present a brief overview of recent efforts in this direction.

\noindent
\textbf{Vision Language Foundation Models.}
The vast amount of easily accessible web-scale image-text pairs has brought about a wave of research innovations in vision language foundation models~\cite{frome2013devise, zhang2022contrastive,wang2016learning, li2021align, li2020oscar, lu2019vilbert,su2019vl}. The pioneering works, CLIP~\cite{radford2021learning} and ALIGN~\cite{jia2021scaling}, are trained with contrastive loss, showing impressive generalization capabilities for downstream tasks by aligning pairs of images and texts in a cross-modal shared embedding space. Subsequent efforts extend the image-text contrastive method to a broader spectrum, such as the image-text-label space proposed in UniCL~\cite{yang2022unified} and a wider range of tasks as well as modalities supported in Florence~\cite{yuan2021florence} and INTERN~\cite{shao2021intern}.
However, contrastive-based methods lack the ability to generate language, which limits their application in open-ended tasks such as captioning or visual question answering.

On the other hand, 
the success of large language models such as GPT series~\cite{radford2018improving,radford2019language,brown2020language,ouyang2022training}, PaLM~\cite{chowdhery2022palm,anil2023palm}, and LLaMA~\cite{touvron2023llama},
has been attracting a lot of research interest~\cite{tsimpoukelli2021multimodal,wang2021simvlm, wang2022omnivl,wang2022git, singh2022flava, li2022grounded,huang2023language} in augmenting the large language models with visual capabilities. Mostly, these models cast a wide range of open-ended vision tasks as text prediction problems, mapping visual input content to language semantics to enable general-purpose visual and language understanding. BEIT3~\cite{wang2022image} unifies the pretraining task in a masked data modeling manner. CoCa~\cite{yu2022coca} and BLIP~\cite{li2022blip,li2023blip} unifies contrastive learning and generative learning. Flamingo~\cite{alayrac2022flamingo} accepts arbitrarily interleaved visual data and text
as input and generates text in an open-ended manner by learning on a broad diversity of vision language tasks.
LLaVA~\cite{liu2023visual} exploits visual instruction tuning by converting image-text pairs into an instruction-following format.
GLIP v2~\cite{zhang2022glipv2} and Kosmos v2~\cite{peng2023kosmos} leverage grounded image-text pairs to further unlock the grounding capability of multimodal large language models.
Our work differs from LLaVA~\cite{liu2023visual} in that, unlike open-ended visual tasks such as visual question answering that can be naturally formulated in an instruction-following format, we attempt to formulate vision tasks, such as segmentation and keypoint detection, into an instruction-following framework. This is challenging due to the unclear instructions and lack of specific guidelines in these tasks.

\noindent
\textbf{Vision Generalist Models.}
Seeking a unified model that, once trained, can be directly used to seamlessly address a wide variety of vision tasks, has been an enduring aspiration in the computer vision community.
Multi-task learning~\cite{zhu2022uni,li2023uni, jaegle2021perceiver, jaegle2021perceiver} has become more and more popular.
The key challenge lies in the diversity and complexity of the various structure of task outputs. 
Currently, there are two major interfaces for output unification: language-like generation and image-resembling generation.
Most existing attempts for vision generalists take inspiration from sequence-to-sequence models in the NLP field and model a sequence of discrete tokens through next token prediction~\cite{wang2022ofa,wang2022git,gupta2022towards,reed2022generalist,chen2021pix2seq}. Pix2Seq v2~\cite{chen2022unified} unifies object detection, instance segmentation, keypoint detection, and
image captioning by quantizing the continuous image coordinates for the first three tasks.
Unified IO~\cite{lu2022unified} further unifies dense structure outputs
such as images, segmentation masks, and depth maps using a vector
quantization variational auto-encoder (VQ-VAE)~\cite{van2017neural}.

As quantization inevitably introduces information loss during discretization, another direction of unification aims to explore the image itself as a natural interface for vision generalists~\cite {bar2022visual,wang2023images}. Painter~\cite{wang2023images} formulates the dense prediction task as a masked image inpainting problem and demonstrates in-context capability in vision tasks such as depth estimation, semantic segmentation, instance segmentation, keypoint detection, and image restoration.
Recently, PromptDiffusion~\cite{wang2023context} also exploits in-context visual learning with a text-guided diffusion model~\cite{rombach2022high} and integrates the learning of six different tasks, i.e., image-to-depth, image-to-HED, image-to-segmentation and vice versa.
Our work also examines image-resembling generation. However, in contrast to in-context learning,
Unlike previous works~\cite{wang2023images,wang2023context} that also explore natural language instructions, our method introduces a more favorable instruction alignment compared to the implicit task intention deducted from in-context learning.
Moreover, with such explicit instructions, we further unify semantic image editing tasks, which are crucial use cases in image-resembling generation.

\section{Method}
\label{sec:method}
We present InstructDiffusion, a novel generalist modeling interface designed for a diverse range of vision tasks. By leveraging the Denoising Diffusion Probabilistic Model (DDPM), we treat all computer vision tasks as human-intuitive image manipulation processes with outputs in a flexible and interactive pixel space.
Several existing multi-modal models, such as Flamingo~\cite{alayrac2022flamingo} and BLIP2~\cite{li2023blip}, inherently produce natural language as their target output, thereby restricting their capabilities to visual question answering and image captioning. In contrast, our approach posits that formulating various vision tasks, including segmentation, keypoint detection, and image synthesis as image-resembling generation processes, is more intuitive, straightforward, and readily assessable for human evaluation.

Our primary focus is on three output formats: 3-channel RGB images, binary masks, and key points. These outputs adequately encompass a wide range of vision tasks, including keypoint detection, semantic segmentation, referring segmentation, semantic image editing, and several image enhancement tasks such as deblurring, denoising, and watermark removal.
We first discuss the essential instructional format design for the vision tasks currently covered in Section~\ref{sec:instructional_image_editing}, followed by an in-depth explanation of the training data preparation to ensure optimal model performance in Section~\ref{sec:training_data_construction}. Lastly, we describe a unified framework with a simple architecture in Section~\ref{sec:unified_framework}.

\subsection{Unified Instructional for Vision Tasks }
\label{sec:instructional_image_editing}


The unified modeling interface for all tasks is referred to as Instructional Image Editing. By denoting the training set as $\{\bm{x}^i\}$, each training data $\bm{x}^i$ can be represented in the form of $\{c^i, s^i, t^i\}$, where $c^i$ signifies the control instruction, while $s^i$ and $t^i$ represent the source and target images, respectively. Within this context, our method aims to generate a target image $t^i$ that adheres to the given instruction $c^i$ when provided with an input source image $s^i$.

In the context of semantic image editing tasks, InstructPix2Pix~\cite{brooks2023instructpix2pix} is a recent representative work that demonstrates a natural fit. For other vision tasks, the challenge involves creating appropriate instructions and subsequently establishing a corresponding target image. Although natural language instruction has been utilized extensively in previous approaches, such as Pix2Seq~\cite{chen2021pix2seq} and UnifiedIO~\cite{lu2022unified}, we contend that terms like "semantic segmentation" or "keypoint detection" are better perceived as indicators rather than instructions. In contrast, our approach involves providing highly detailed instructions, enabling the model to comprehend the instructions rather than merely model a fixed bias based on the indicator.


\noindent\textbf{Keypoint detection.}
It endeavors to precisely locate key object components within an image, such as the left eye of a face, the right shoulder of an individual, or the nose of a dog. Traditionally, heatmap regression has served as the standard learning approach, where ground truth heatmaps are generated by overlaying 2D Gaussian kernels on all keypoints. In contrast, this work introduces a more natural and easily assessable output by providing extensively detailed instructions, thereby enhancing the overall process of keypoint detection in various applications.
An exemplary instruction might be, \emph{"Please use red to encircle the left shoulder of the man."} In this instance, the output image should exhibit a red circle at the corresponding location (\ie, the left shoulder of the man in the image), while the rest of the region remains unaltered. This innovative approach facilitates a more intuitive comprehension of the keypoint detection process while simultaneously refining the model's capacity to understand the meaning of different object components.

\noindent\textbf{Segmentation.}
For semantic and referring segmentation, the objective is to identify the region of a particular object within the input image. An illustrative example of this instruction would be \emph{"apply a blue semi-transparent mask to the rightmost dog while maintaining the remainder unaltered."} Consequently, the resulting image is determined and features a blue mask on the appropriate dog. We require the mask to be semi-transparent instead of opaque, thereby facilitating the human evaluation of the predicted mask's accuracy. Moreover, our experiments indicate that the semi-transparent mask also augments the segmentation performance.

\noindent\textbf{Image enhancement and image editing.}
Image enhancement such as deblurring, denoising, and watermark removal inherently yields output images, and the same applies to image editing. Consequently, we only need to construct instructions which shall clearly specify the operation to be performed. Detailed examples include \emph{``Make the image much sharper"} for image deblurring, \emph{``Please remove the watermark on the image"} for watermark removal, and \emph{``add an apple in the woman's hand"} for image editing.

To enhance the diversity of instructions, we first manually write 10 instructions for each task. Then we use GPT-4 to rewrite and expand the diversity of these instructions, thereby mimicking user input to the system. Subsequently, one instruction is chosen at random during the training process. This approach, which incorporates diverse and intuitive instructions, has been observed to substantially augment the model's multi-task fusion capabilities.


\subsection{Training Data Construction}
\label{sec:training_data_construction}
As a proof-of-concept, we focus on investigating whether different tasks benefit each other under such image-resembling unification, instead of scaling data as much as possible for optimal performance at the extreme limits. We adopt widely used publicly available datasets and construct the ground truth target image according to the instruction template. 
For example, we use COCO-Stuff~\cite{caesar2018cvpr} for semantic segmentation and use COCO~\cite{lin2014microsoft}, MPII~\cite{Andriluka14mpii}, CrowPose~\cite{li2018crowdpose} and AIC~\cite{wu2017aic} for keypoint detection.  
More details will be presented in Sec~\ref{sec:settings}.

For image editing, InstructPix2Pix (IP2P)~\cite{brooks2023instructpix2pix} pioneered the use of a synthetic training dataset by leveraging GPT-3~\cite{brown2020language} for generating instructions and Prompt2Prompt~\cite{hertz2022prompt} for creating output images. However, the synthesized source and target images exhibit varying quality and non-negligible artifacts, with most instructions focusing on global style modifications rather than local alterations.
Furthermore, MagicBrush~\cite{zhang2023magicbrush} introduced a dataset comprising over 10,000 manually annotated triples, but its size is limited when compared to other vision tasks. Consequently, in addition to existing datasets such as IP2P~\cite{brooks2023instructpix2pix}, GIER~\cite{shi2020gierbenchmark}, GQA~\cite{yildirim2023inst}, and MagicBrush~\cite{zhang2023magicbrush}, we propose a novel dataset called Image Editing in the Wild (IEIW), which encompasses 159,000 image editing pairs that cover a wide range of semantic entities and diverse levels of semantic granularity.
To expand the scope of image editing data, we assemble the IEIW dataset by drawing from the following three distinct resources:
\noindent \textbf{Object removal.} Object removal is a very common type of image editing. Inspired by Inst-Inpaint~\cite{yildirim2023inst}, we use the referring segmentation dataset PhraseCut~\cite{wu2020phrasecut} to construct the instructional object removal data. PhraseCut offers images with referring phrases for corresponding regions. We set these regions as a mask and use LAMA~\cite{suvorov2022resolution} to inpaint them, transforming them into instructional inpainting datasets. Notably, we also swap input and output images, and reverse the instructions like "remove the blue bird on top of the tree" to "add a blue bird on top of the tree" to further supplement data from the perspective of adding components.


\noindent\textbf{Object replacement.} We propose a data construction pipeline for generating training data that targets the scenario of substituting certain specific objects, which is another
essential feature for image editing.
To automatically generate training triplets, we rely on SA-1B~\cite{kirillov2023segment} and OpenImages~\cite{kuznetsova2020open} datasets, which provide multiple regions in an image with semantic meaning.
Specifically, we first build a gallery database consisting of diverse image patches based on those semantic-aware regions. Given a source image from OpenImages or SA-1B, we randomly select a semantic region, which is used as a query patch to retrieve its nearest neighbors from the aforementioned constructed gallery database. The retrieved similar patches are regarded as reference images to the source image, both of which are fed to PaintByExample~\cite{yang2023paint} for generating a target image. In this way, we obtain the source image as well as the modified target image. To produce instruction, we utilize an image captioning tool, such as BLIP2~\cite{li2023blip}, to yield the source caption as well as the target caption, and then generate a possible instruction through a large language model.
For example, given the captions ``a running dog" and ``a cute cat with black and white stripes", a possible instruction is ``please change the running dog to a cute cat with black and white stripes".
We can generate quite an amount of paired data for training using this construction pipeline. 

\noindent\textbf{Web crawl.} In order to achieve greater alignment with authentic user needs and enhance the overall user experience, we gather genuine user requests along with the corresponding outcomes delivered by seasoned Photoshop professionals sourced from the website. To ensure the accuracy and relevance of the data, we search in Google by utilizing the keyword "photoshop request". This approach enables us to amass a substantial dataset comprising over 23,000 data triplets, which further aids in refining our understanding of user requirements and reduces the domain gap between training and inference.

In order to guarantee the quality of the training data, we further utilize image quality assessment tools to eliminate substandard data. Specifically, we apply Aesthetics Score and GIQA~\cite{gu2020giqa} as image quality evaluation metrics, specifically utilizing LAION-Aesthetics-Predictor~\cite{schuhmann2022laion} for Aesthetics Score and constructing a KNN-GIQA model on LAION-600M~\cite{schuhmann2022laion} images for calculating GIQA scores. We exclude two categories of data: i) target images with low-quality scores, and ii) a significant discrepancy in quality scores between the source image and its corresponding target image. Our findings indicate that this data-filtering process is of vital importance.

\begin{figure}
    \centering
    \captionsetup{type=figure}
    \includegraphics[width=0.48\textwidth, trim={150 210 150 210},clip]{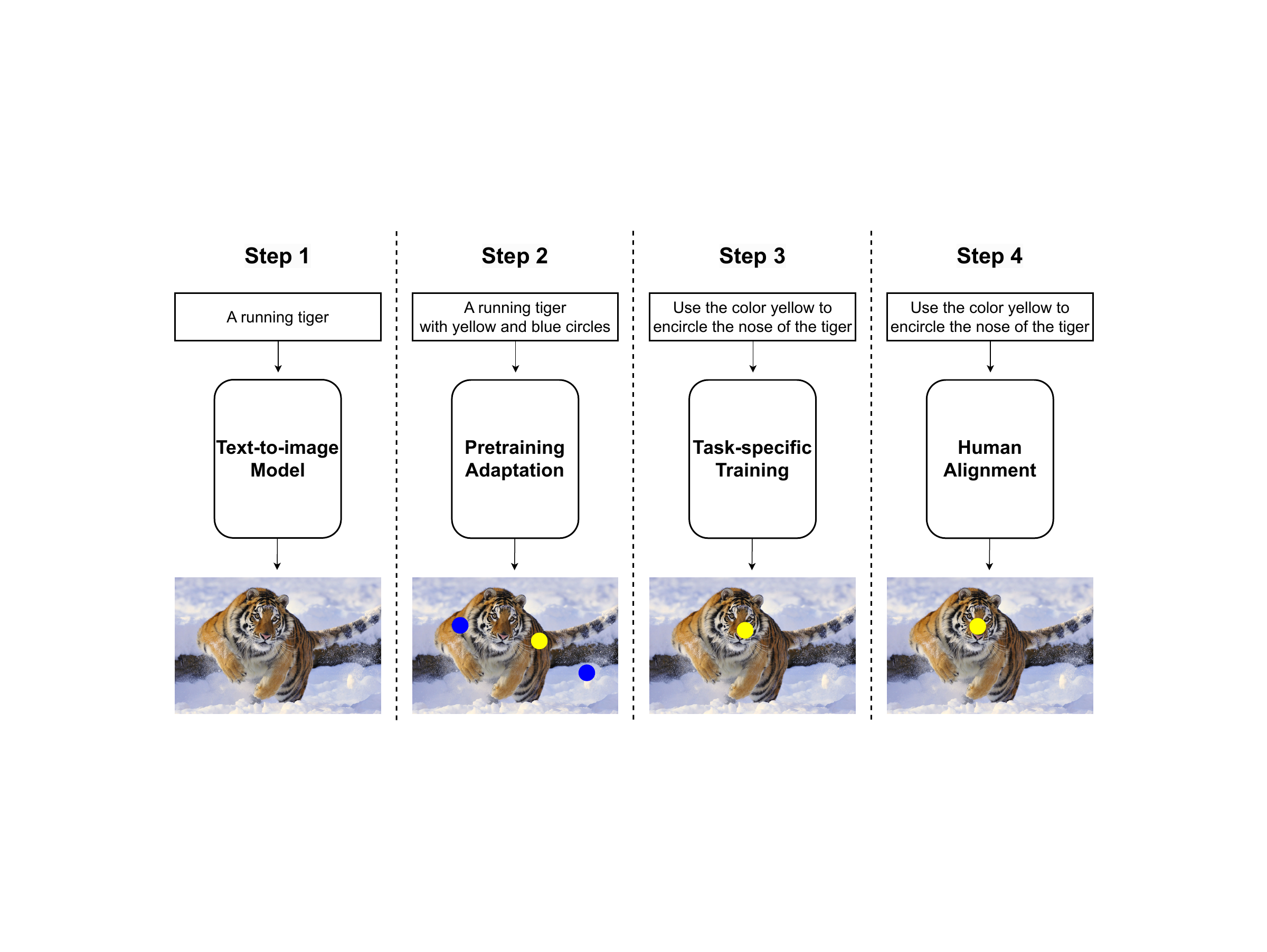}
    \vspace{-1.5em}
    \captionof{figure}{Training pipeline of our method. To illustrate concisely, we take keypoint detection as an example.}
    \label{fig:pipeline}
    \vspace{-1.3em}
\end{figure}

\subsection{Unified Framework}
\label{sec:unified_framework}
Our framework is based on diffusion, as diffusion models have experienced significant success in modeling complex image distributions. As illustrated in Figure~\ref{fig:pipeline}, our training procedure comprises three stages: pretraining adaptation, task-specific training, and instruction tuning.

\noindent \textbf{Pretraining adaptation.}
Stable Diffusion (SD)~\cite{rombach2022high} is recognized as one of the most robust open-source text-to-image models currently accessible, prompting our decision to utilize Stable Diffusion v1.5 as the foundation for our work. Initially, stable diffusion operates as a mapping mechanism that converts textual captions into natural images. However, our desired images might encompass segmentation masks or keypoint indicators, which substantially deviate from typical natural images. Consequently, our preliminary phase involves fine-tuning the stable diffusion model and adjusting the diffusion output distribution.

Since we require diffusion models to be capable of generating images ``with a foreground mask" or ``with some special mark", we employ existing segmentation or keypoint detection datasets to produce such data. The remaining challenge lies in the development of suitable captions that accurately depict these images while maintaining the intrinsic text-to-image generation capability. This is achieved by augmenting the original image caption with a suffix, such as "with a few different color patches here and there" or "surrounded with a red circle." By fine-tuning the diffusion model with these modified image captions, we can theoretically empower the model to generate any images within the desired output domain.

\renewcommand{\arraystretch}{1.25}
\begin{table*}[]
\setlength\tabcolsep{11.25pt}
\centering
\caption{The number of effective training samples used for different tasks.}
\vspace{-0.5em}
\begin{tabular}{c|cccc}
\bottomrule
Task & Keypoint Detection & Segmentation & Image Enhancement & Image Editing \\ 
\hline
\# Effective training samples & 245k & 239k & 46k & 425k \\
 \toprule
\end{tabular}
\label{tab:training_samples}
 \vspace{-1.0em}
\end{table*}

\noindent \textbf{Task-specific training.}
In the second stage, our goal is to further fine-tune the diffusion model, enhancing its comprehension of various instructions for different tasks. We follow InstructPix2Pix~\cite{brooks2023instructpix2pix} and inject source images by concatenating them with the noise input, subsequently expanding the input channels of the first layer. We train our model using all data containing various tasks. Since the amount of data for each task is quite different, in order to maintain a balance, we manually set different sampling weights for different databases. The number of effective training samples used for different tasks is shown in Table~\ref{tab:training_samples}. For a data triplet ${s_i, c_i, t_i}$, the diffusion process adds noise to the encoded latent $z = \mathcal{E}(t_i)$ producing a noisy latent $z_t$. We fine-tune the diffusion network $\epsilon_\theta$ by minimizing the following latent diffusion objective:
\begin{equation}
L = \mathbb{E}_{(s_i, c_i, t_i) \sim \mathcal{P}(x), \epsilon \sim \mathcal{N}(0, 1), t }\Big[ \Vert \epsilon - \epsilon_\theta(z_{t}, t, s_i, t_i) \Vert_{2}^{2}\Big]
\label{eq:loss}
\end{equation}


\noindent \textbf{Human alignment.}
To further improve the quality of editing, we have followed the idea of instruction tuning~\cite{wei2021flan} from Large Language Models. In LLM literature, instruction tuning~\cite{wei2021flan} is used to teach the model to solve a task following the instruction. However, we conduct instruction tuning differently from that in LLM. For each sample in the benchmark, we generate different editing results using $20$ different sampling classifier-free guidance~\cite{ho2022classifier}. Then, we ask subjects to select the best 0-2 edited images to formulate the instruction-tuning dataset. The whole dataset contains $1,000$ images. We use this dataset to further fine-tune our model for about 10 epochs.

\section{Experiments}


\subsection{Settings}
\label{sec:settings}

\noindent 
\textbf{Training samples.}
Our model is trained on samples consisting of \{instruction, source image, target image\}, encompassing the aforementioned vision tasks, \ie, keypoint detection, semantic segmentation, referring segmentation, image enhancement including denoising, deblurring and watermark removal, and image editing.
Specifically for \textbf{keypoint detection}, we adopt four classical datasets, namely COCO~\cite{lin2014microsoft} containing $149$K images with each labeled $17$ keypoints, CrowdPose~\cite{li2018crowdpose} consisting of $35$K images each with $14$ keypoints, MPII~\cite{Andriluka14mpii} with $22$K images labeled with $16$ keypoints, and AIC~\cite{wu2017aic} including $378$K images annotated with $14$ keypoints.
Throughout our training process, for each image, we employ a random selection of between 1 and 5 keypoints, and assign these keypoints with random colors. Accordingly, the instruction is produced through templates filled with the class of keypoints and the specific color, and the target image is generated by positioning small circles on the chosen keypoints, each circle taking on the color corresponding to its respective keypoint.
For \textbf{segmentation}, we select COCO-Stuff~\cite{caesar2018cvpr} as semantic segmentation training dataset while gRefCOCO~\cite{GRES} and RefCOCO~\cite{yu2016modeling} as referring segmentation training dataset.
We collect a series of prompt templates with the help of large language models to serve as text instructions. An example is ``place a {color} mask on {object}." During training, we randomly select a color for ``{color}" and replace ``{object}" with the corresponding category name in semantic segmentation or referring in referring segmentation. The target image is generated by placing a mask using its corresponding color with a transparency of 0.5 over the object.
For \textbf{image enhancement}, we focus on three tasks: deblurring, denoising, and watermark removal. For these tasks, we utilize the GoPro~\cite{Nah_2017_CVPR} containing 2103 images and REDS~\cite{Nah_2019_CVPR_Workshops_REDS} dataset with 24,000 images for deblurring, the SIDD~\cite{SIDD_2018_CVPR} dataset composed of 320 images for denoising, and the CLWD~\cite{Liu_2021_WACV} dataset containing 60,000 images for watermark removal. 
Lastly for \textbf{image editing}, as mentioned in Sec.~\ref{sec:training_data_construction}, we adopt 7 editing datasets, including filtered InstructPix2Pix~\cite{brooks2023instructpix2pix} dataset containing 561K samples, 8K samples in MagicBrush~\cite{zhang2023magicbrush} training dataset, GIER~\cite{shi2020gierbenchmark} with 5K samples, GQA~\cite{yildirim2023inst} inpainting dataset with 131K samples, VGPhraseCut~\cite{wu2020phrasecut} composed of 85K samples, our generated dataset with 51K produced samples, and an internal dataset representing real editing scenario, which contains 23K training triplets. 
To ensure balanced sampling across each task, we implemented distinct sampling weights due to the considerable variance in the number of training images across different datasets. Table~\ref{tab:training_samples} illustrates the number of effective training samples we used in our framework.

\noindent
\textbf{Implementation details.}
We utilize Stable Diffusion~\cite{rombach2022high} v1.5 as initialization to leverage a text-to-image generation prior.
The input image resolution is preprocessed
to $256 \times 256$, and the learning rate is fixed to $1 \times 10^{-4}$ during training. In addition, we adopt an EMA rate of 0.9999 to stabilize the training. Our model is trained using a batch size of 3072 for a total of 200 epochs, which requires approximately 4 days of computation on 48 NVIDIA V100 GPUs. Once trained, our model is readily applicable and can be directly used for different vision tasks. For each task, we provide a comprehensive comparison and an in-depth analysis of its performance in the subsequent sections. During the human alignment stage, we use an EMA rate of 0.99 to help the model quickly adapt to the instruction-tuning dataset.

\renewcommand{\arraystretch}{1.1}
\begin{table}[t]
		\caption{Average precision comparison on the COCO val2017, HumanArt and AP-10K datasets. We evaluate the official large models of the competitors to ensure fairness. The ground truth bounding boxes are used for all results. The best-performing generalist models are highlighted in bold.
		}
		\vspace{-0.25cm}
		\centering\setlength{\tabcolsep}{6.75pt}
		\label{table:coco}
		\begin{tabular}{l|c|c|c}
			\bottomrule
		    Method & COCO val & HumanArt & AP-10K \\\hline
                \multicolumn{4}{c}{{\fontsize{8pt}{18pt}\selectfont Specialized Models}} \\ 
			\hline
			{PCT~\cite{Geng23PCT}} & \textcolor{gray}{80.2} & \textcolor{gray}{63.7} & \textcolor{gray}{14.6} \\
			{ViTPose~\cite{xu2022vitpose}} & \textcolor{gray}{82.0} & \textcolor{gray}{64.1} & \textcolor{gray}{14.7} \\\hline
                \multicolumn{4}{c}{\fontsize{8pt}{18pt}\selectfont Generalist Models} \\ 
                \hline
			Unified-IO~\cite{lu2022unified} & 25.0 & 15.7 & 7.6 \\
		      Painter~\cite{wang2023images} & 70.2 & 12.4 & 15.3 \\
			Ours & \textbf{71.2} & \textbf{51.4} & \textbf{15.9} \\
			\toprule
		\end{tabular}
		\vspace{-.5cm}
\end{table}

\begin{figure*}
    \centering
    \captionsetup{type=figure}
    (a) \includegraphics[height=0.158\textwidth, trim={5 0 25 0},clip]{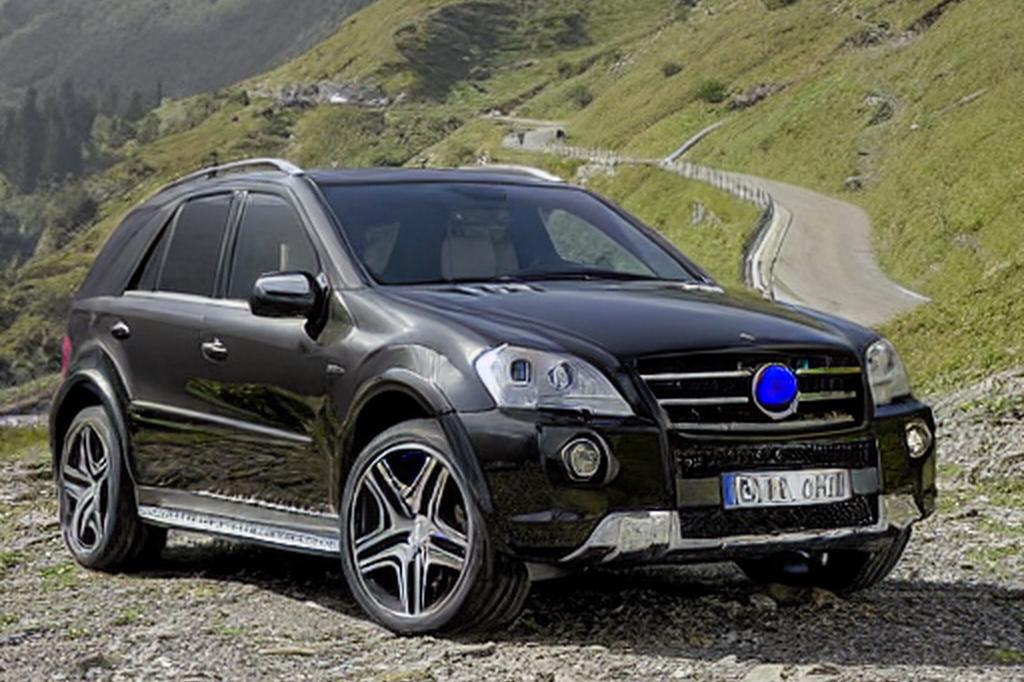}~
    (b) \includegraphics[height=0.158\textwidth, trim={0 0 10 0},clip]{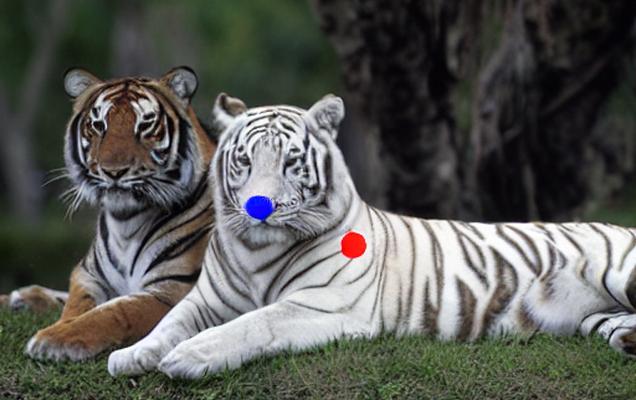}~
    (c) \includegraphics[height=0.158\textwidth, trim={40 0 60 0},clip]{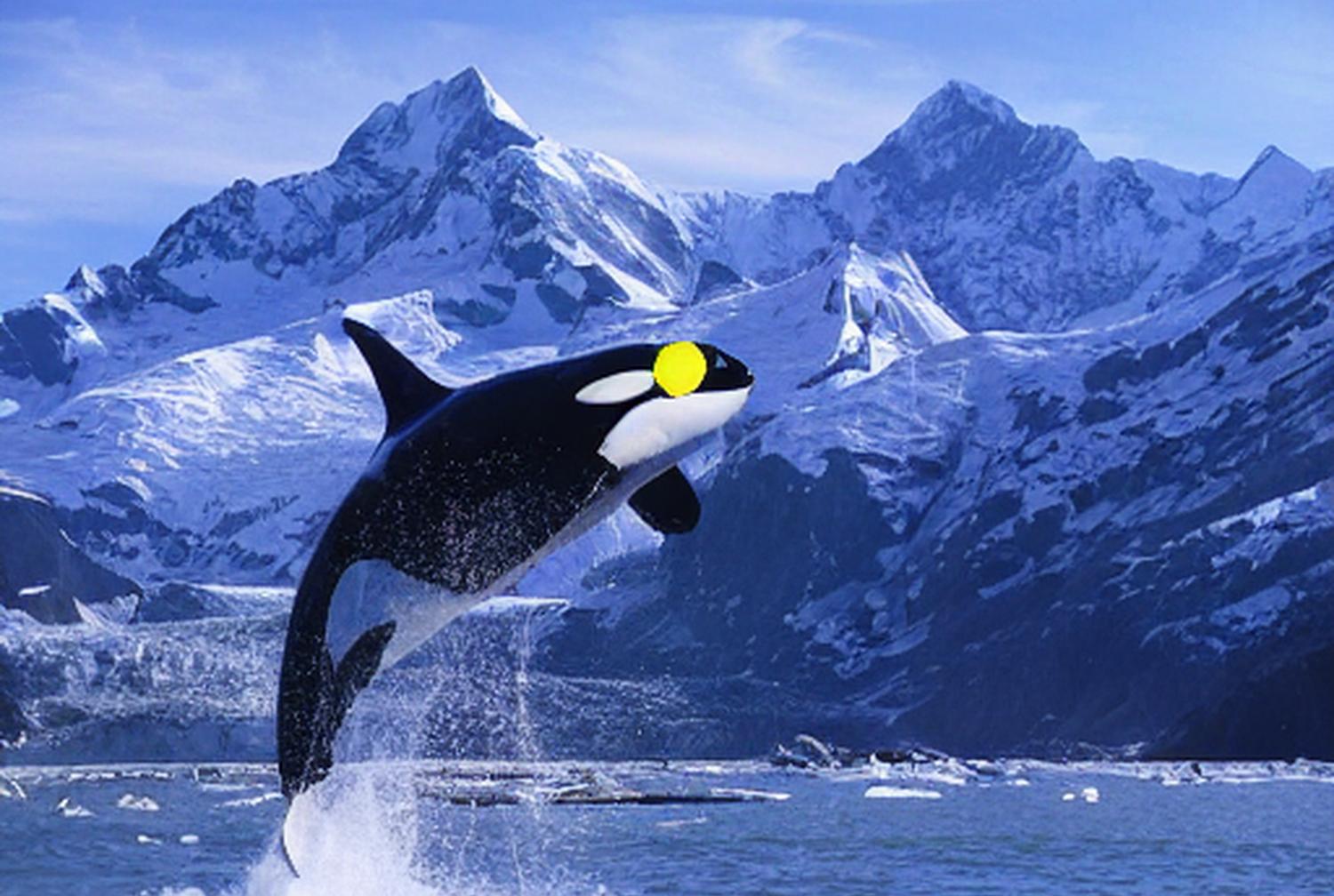}~
    (d) \includegraphics[height=0.158\textwidth]{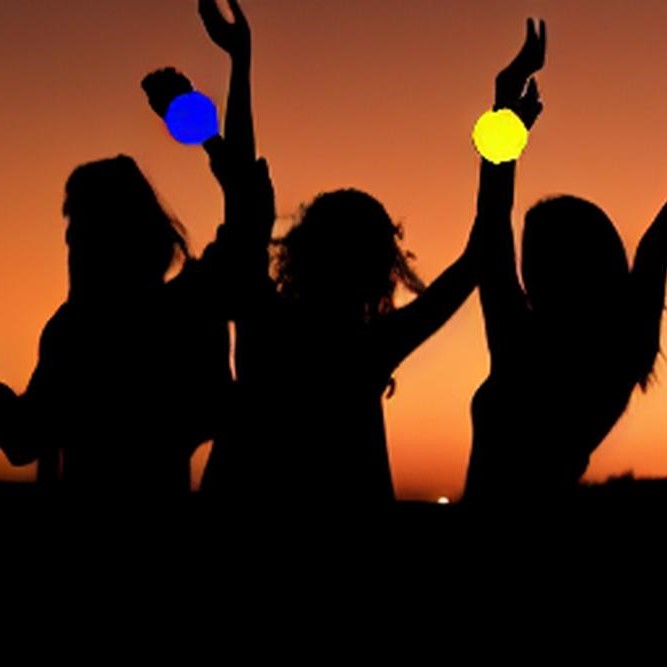}~\\
    \vspace{-0.3em}
    \caption{The keypoint detection results generated by our model. The instructions are as follows: (a) Mark the \textbf{car logo} with a \textbf{blue} circle. (b) Put a \textbf{blue} circle on the \textbf{nose} of the \textbf{white tiger} and use the \textbf{red} color to draw a circle around the \textbf{left shoulder} of the \textbf{white tiger}. (c) Create a \textbf{yellow} circle around the \textbf{right eye} of the \textbf{whale}.  (d) Use the color \textbf{blue} to encircle the \textbf{right wrist} of the \textbf{person on the far left} and draw a \textbf{yellow} circle over the \textbf{left wrist} of the \textbf{person on the far right}.}~
    \vspace{-0.8em}
    \label{fig:keypoint}
\end{figure*}%

\begin{figure*}
    \centering
    \captionsetup{type=figure}
    (a) \includegraphics[height=0.158\textwidth, trim={35 0 0 0},clip]{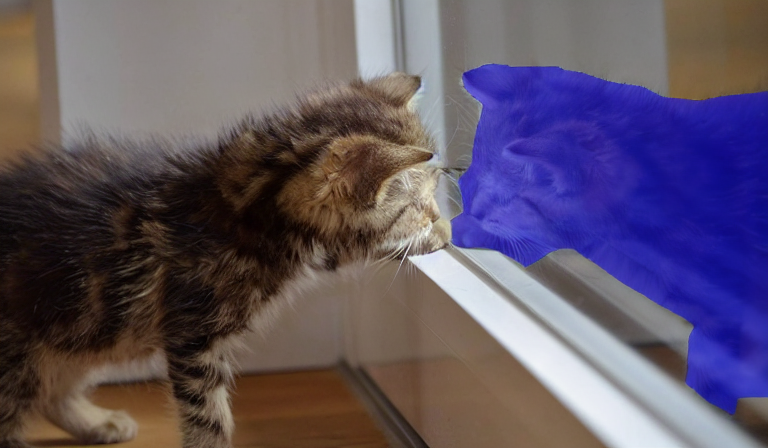}~
    (b) \includegraphics[height=0.158\textwidth, trim={0 0 20 40},clip]{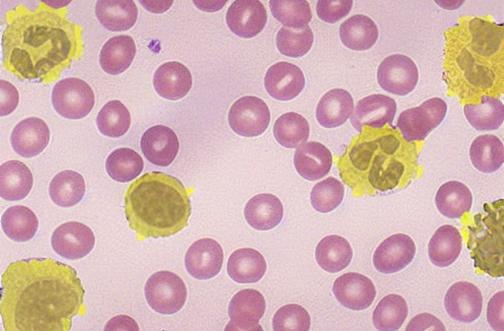}~
    (c) \includegraphics[height=0.158\textwidth, trim={0 0 0 10},clip]{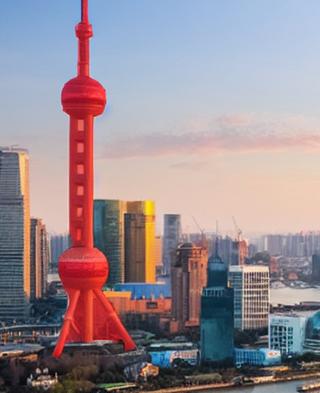}~
    (d) \includegraphics[height=0.158\textwidth, trim={0 0 0 44},clip]{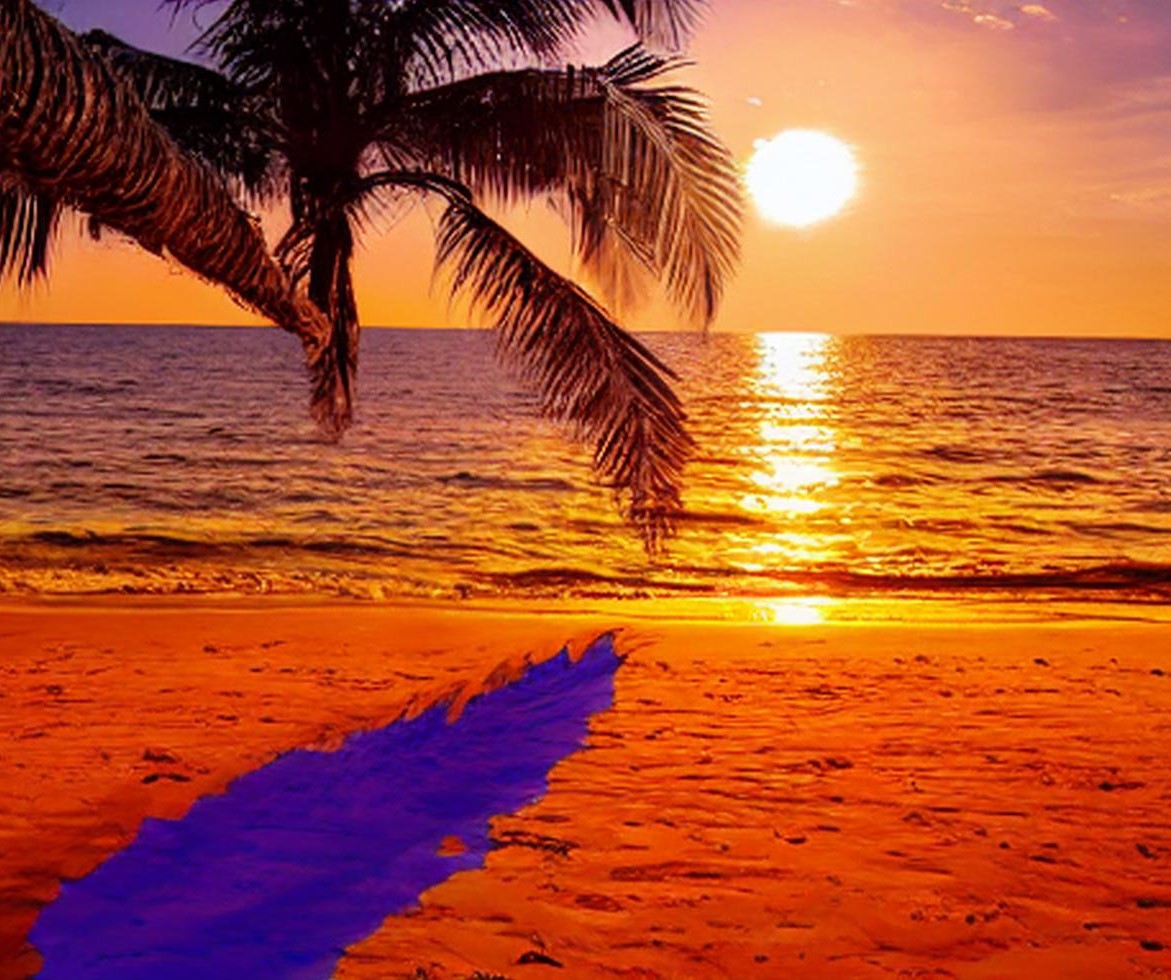}~\\
    \vspace{-0.3em}
    \caption{The segmentation results generated by our model. The instructions are as follows: (a) Mark the pixels of \textbf{cat in the mirror} to \textbf{blue} and leave the rest unchanged. (b) Fill in the pixels of \textbf{neutrophil} with \textbf{yellow}, retaining the existing colors of the remaining pixels. (c) Modify the pixels of \textbf{Oriental Pearl Tower} to \textbf{red} without affecting any other pixels. (d) Paint the pixels of \textbf{shadow} in \textbf{blue} and maintain the current appearance of the other pixels. 
    }~
    \vspace{-1.9em}
    \label{fig:refseg}
\end{figure*}%

\renewcommand{\arraystretch}{1.15}
\begin{table*}[]
\setlength\tabcolsep{2.85pt}
\centering
\caption{Quantitative results on referring segmentation in terms of cIoU. U: UMD split. G: Google split. The best-performing generalist models are highlighted in bold.}
\vspace{-0.8em}
\begin{tabular}{l|c|ccc|ccc|ccc|cccc}
\bottomrule
\multirow{2}{*}{Method} & \multicolumn{1}{c|}{gRefCOCO} & \multicolumn{3}{c|}{RefCOCO} & \multicolumn{3}{c|}{RefCOCO+} & \multicolumn{3}{c|}{G-Ref} & \multicolumn{4}{c}{RefClef} \\
 & val & val & test A & test B & val & test A & test B & val$_\text{(U)}$ & test$_\text{(U)}$ & val$_\text{(G)}$ & val & testA & testB & testC\\ \hline
\multicolumn{15}{c}{{\fontsize{8pt}{18pt}\selectfont Specialized Models}} \\ \hline 
{LAVT~\cite{yang2022lavt}} & \textcolor{gray}{57.64} & \textcolor{gray}{72.73} & \textcolor{gray}{75.82} & \textcolor{gray}{68.79} & \textcolor{gray}{56.86} & \textcolor{gray}{62.29} & \textcolor{gray}{48.14} & \textcolor{gray}{58.65} & \textcolor{gray}{59.17} & \textcolor{gray}{61.16} & \textcolor{gray}{21.22} & \textcolor{gray}{44.77} & \textcolor{gray}{24.78} & \textcolor{gray}{47.08}\\
{ReLA~\cite{GRES}} & \textcolor{gray}{62.42} & \textcolor{gray}{73.21} & \textcolor{gray}{75.24} & \textcolor{gray}{68.72} & \textcolor{gray}{56.10} & \textcolor{gray}{62.26} & \textcolor{gray}{47.89} & \textcolor{gray}{59.71} & \textcolor{gray}{60.40} & \textcolor{gray}{61.37} & \textcolor{gray}{20.68} & \textcolor{gray}{43.08} & \textcolor{gray}{22.57} & \textcolor{gray}{45.94} \\ \hline
\multicolumn{15}{c}{\fontsize{8pt}{18pt}\selectfont Generalist Models} \\ \hline 
Unified-IO~\cite{lu2022unified} & 17.31 & 46.42 & 46.06 & 48.05 & 40.50 & 42.17 & \textbf{40.15} & 48.74 & 49.13 & 44.30 & 40.13 & 43.33 
& 48.07 & 32.47 \\ 
    Ours & \textbf{67.36} & \textbf{61.74} & \textbf{65.20} & \textbf{60.17} & \textbf{46.57} & \textbf{52.32} & 39.04 & \textbf{51.17} & \textbf{52.02} & \textbf{52.18} & \textbf{49.58} & \textbf{54.73} & \textbf{54.82} & \textbf{40.34}\\ \toprule
\end{tabular}
\label{tab:quantitative_ref_seg}
\vspace{-0.5em}
\end{table*}

\renewcommand{\arraystretch}{1.1}
\begin{table*}[]
\setlength\tabcolsep{14.4pt}
\centering
\caption{Quantitative results on semantic segmentation in terms of mcIoU. The best-performing generalist models are highlighted in bold.}
\vspace{-0.8em}
\begin{tabular}{l|cccccc}
\bottomrule
Method & ADE-847 & PC-459 & ADE-150 & PC-59 & VOC & COCO-Stuff \\ \hline
\multicolumn{7}{c}{{\fontsize{8pt}{18pt}\selectfont Specialized Models}} \\ \hline  
{SimSeg~\cite{liang2023open}} & \textcolor{gray}{10.43} & \textcolor{gray}{13.98} & \textcolor{gray}{25.89} & \textcolor{gray}{53.55} & \textcolor{gray}{39.27} & \textcolor{gray}{40.26} \\
{OvSeg~\cite{xu2021}} & \textcolor{gray}{13.85} & \textcolor{gray}{22.72} & \textcolor{gray}{36.50} & \textcolor{gray}{60.93} & \textcolor{gray}{38.50} & \textcolor{gray}{48.76} \\
{SAN~\cite{xu2023side}} & \textcolor{gray}{18.84} & \textcolor{gray}{33.32} & \textcolor{gray}{38.79} & \textcolor{gray}{63.31} & \textcolor{gray}{46.14} & \textcolor{gray}{50.15} \\ \hline
\multicolumn{7}{c}{\fontsize{8pt}{18pt}\selectfont Generalist Models} \\ \hline 
Painter~\cite{wang2023images} & 5.00 & 8.68 & \textbf{54.50} & 33.67 & 4.67 & 11.91 \\
PromptDiffusion~\cite{wang2023context} & 0.99 & 2.19 & 8.97 & 13.07 & 11.69 & 2.71 \\
Unified-IO~\cite{lu2022unified} & 8.96 & 13.69 & 15.65 & 27.21 & 31.46 & 22.52 \\ 
Ours & \textbf{19.68} & \textbf{28.29} & 33.62 & \textbf{59.00} & \textbf{72.55} & \textbf{53.17} \\ 
\toprule
\end{tabular}
\label{tab:quantitative_sem_seg}
\vspace{-1.2em}
\end{table*}

\begin{figure*}
    \centering
    \captionsetup{type=figure}
    \includegraphics[width=0.997\textwidth]{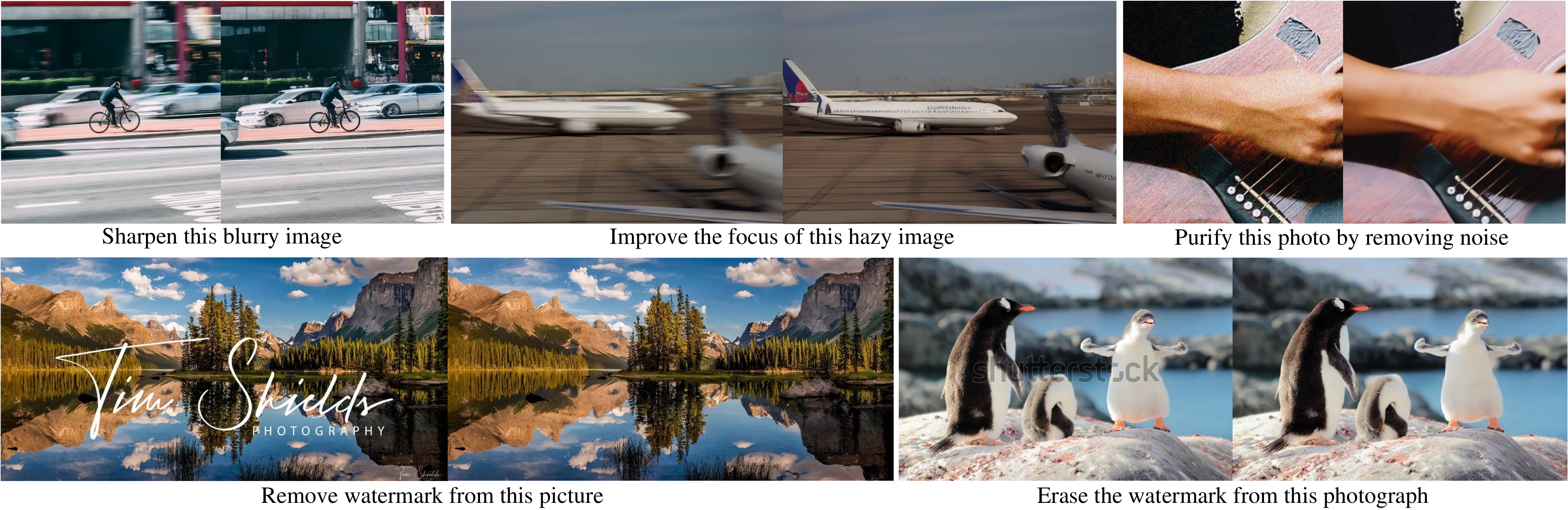}
    \vspace{-1.0em}
    \captionof{figure}{InstructDiffusion is also applicable to low-level vision tasks, including image deblurring, denoising, and watermark removal.}
    \label{fig:lowlevel}
    \vspace{-0.2em}
\end{figure*}%

\renewcommand{\arraystretch}{1.1}
\begin{table*}[]  
\setlength\tabcolsep{7.52pt}  
\centering  
\caption{Quantitative results on image editing and image enhancement. For editing tasks (Replace, remove, and add), the results are CLIP-Sim/AP score. For enhancement tasks, the number reflects the PSNR metric. The numbers in parentheses indicate the results obtained by reconstructing the ground truth images using VAE, representing the performance upper bound achievable with the used VAE model.}  
\vspace{-0.5em}
\label{tab:editing-and-lowlevel}
\begin{tabular}{l|ccc|ccc} 
\bottomrule
{} & \multicolumn{3}{c|}{Editing (CLIP-Sim$\uparrow$ / AP Score$\uparrow$)} & \multicolumn{3}{c}{LowLevel (PSNR$\uparrow$)} \\
\hline  
Method & Replace  & Remove  & Add  & Deblur   & Denoise   & Watermark remove \\ \hline 
\multicolumn{7}{c}{\fontsize{8pt}{18pt}\selectfont Specialized Models} \\ \hline  
NAFNet~\cite{chen2022simple} & \textcolor{gray}{-} & \textcolor{gray}{-} & \textcolor{gray}{-} & \textcolor{gray}{33.71} &  \textcolor{gray}{40.30} & \textcolor{gray}{-} \\
WDNet~\cite{Liu_2021_WACV} & \textcolor{gray}{-} & \textcolor{gray}{-} & \textcolor{gray}{-} & \textcolor{gray}{-} & \textcolor{gray}{-} & \textcolor{gray}{40.24}
\\ 
Null-text~\cite{mokady2023null} & \textcolor{gray}{30.71/5.04} & \textcolor{gray}{29.69/4.80} & \textcolor{gray}{30.14/4.95} & \textcolor{gray}{24.52} & \textcolor{gray}{23.29} & \textcolor{gray}{18.31}  
\\
InstructPix2Pix~\cite{brooks2023instructpix2pix} & \textcolor{gray}{29.61/4.99} & \textcolor{gray}{28.82/4.69} & \textcolor{gray}{30.11/4.94} & \textcolor{gray}{22.71} &  \textcolor{gray}{15.14} & \textcolor{gray}{14.96}   \\ 
MagicBrush~\cite{zhang2023magicbrush} & \textcolor{gray}{30.50/4.94} & \textcolor{gray}{29.07/4.67} & \textcolor{gray}{30.69/4.90} & \textcolor{gray}{22.64} & \textcolor{gray}{16.10} & \textcolor{gray}{15.46}  \\  
EDICT~\cite{wallace2023edict} & \textcolor{gray}{29.91/4.91} & \textcolor{gray}{29.33/4.80} & \textcolor{gray}{30.19/4.93} & \textcolor{gray}{24.16} & \textcolor{gray}{24.48}
 & \textcolor{gray}{19.88}  \\
 \hline
\multicolumn{7}{c}{{\fontsize{8pt}{18pt}\selectfont Generalist Models}} \\ \hline  
Painter~\cite{wang2023images} & - & - & - & - &  38.66 & -  \\  

Ours & 30.19/4.90 & 28.88/4.65 & 30.39/4.87 & 23.58 (29.54) & 34.26 (36.56) & 23.60 (24.80)  \\
\toprule
\end{tabular}  
\vspace{-1.2em}
\end{table*} 

\subsection{Keypoint Detection}
We evaluate our model on both the close-set scenario using the COCO validation dataset 
as well as the open-set generalization capability over the unseen dataset: 
HumanArt dataset~\cite{ju2023humanart} and AP-10K animal dataset~\cite{yu21ap10k}. The HumanArt dataset is an artificial human dataset comprised of various forms such as cartoons, shadow plays, and murals, exhibiting a distinct data distribution compared to COCO dataset. The AP-10K Animal dataset is a collection of annotated animal keypoints, which effectively highlights the ability of our model to handle animal keypoints despite being trained only on human keypoint datasets. To enable a more detailed and thorough evaluation, it is essential to extract accurate pose coordinate information, namely precise horizontal and vertical coordinates, rather than simply marking the location with a distinct symbol. To achieve this, we employ a lightweight U-Net structure that post-processes the output image to generate a multi-channel heatmap.
We employ the standard $AP$ (average precision) based on the OKS as our evaluation metrics. Additionally, we utilize the ground truth bounding boxes for all results. Notably, for the AP-10K animal dataset, in order to facilitate comparison with other methods, the OKS is calculated exclusively on the keypoints that overlap with the COCO annotated joints. However, it should be noted that our model possesses the capability to detect keypoints beyond the confines of the training dataset. 

The results of the keypoint detection are presented in Table~\ref{table:coco}. Our approach outperforms other generalist models, Unified-IO~\cite{lu2022unified} and Painter~\cite{wang2023images}, across all evaluated datasets. Particularly we demonstrate a significantly higher level of performance over HumanArt and AP-10K, indicating the powerful generalization ability of our framework.
In comparison to methods specifically designed for keypoint detection, our unified model does not exceed their performance due to localization accuracy limitations. However, it showcases exceptional performance on the entirely unseen animal keypoints dataset, AP-10K. Figure~\ref{fig:keypoint} (a-c) display our results for car and animals keypoint detection. Our model can accurately detect the logo of the car and the keypoints of animals that have never appeared in the keypoint detection training dataset. Figure~\ref{fig:keypoint} (d) demonstrate our capability for referring keypoint detection, showcasing our versatile detection abilities.


\subsection{Segmentation}
Our primary focus lies in assessing the open-vocabulary capability of our model, particularly when evaluating images that contain unseen classes not present during the training phase. Therefore,
besides the COCO-stuff~\cite{caesar2018cvpr}, gRefCOCO~\cite{GRES} and RefCOCO~\cite{yu2016modeling} datasets, 
we conduct evaluation over additional eight datasets, \ie, RefCOCO+~\cite{yu2016modeling}, G-Ref~\cite{mao2016generation}, RefClef~\cite{kazemzadeh2014referitgame} for referring segmentation, and ADE20K-150~\cite{zhou2017scene}, ADE20K-847~\cite{zhou2017scene}, Pascal Context-59~\cite{mottaghi2014role}, Pascal Context-459~\cite{mottaghi2014role}, Pascal VOC~\cite{everingham2012pascal} for semantic segmentation. 
Similar to keypoint detection,
we employ a lightweight U-Net structure that post-processes the output image to extract the binary mask of each individual object.
Adhering to the prevailing convention~\cite{GRES}, we adopt cumulative IoU (cIoU) to measure the performance for referring segmentation.
On the other hand, our approach involves predicting a mask for each semantic category individually. As a result, semantic segmentation can also be perceived as referring to segmentation based on semantic categories.
Thus, we choose to utilize the mean of class-wise cumulative intersection over
union (mcIoU) to quantify the performance of semantic segmentation.

Table~\ref{tab:quantitative_ref_seg} reports the results for referring segmentation. To the best of our knowledge, Unified-IO~\cite{chen2022unified} stands as the sole generalist model with the capability to perform referring segmentation.
It can be seen that our model largely outperforms Unified-IO across almost all datasets.
We also present methods that are specially designed for referring segmentation. Interestingly, our approach achieves an unexpectedly significant improvement over the RefClef dataset.
Table~\ref{tab:quantitative_sem_seg} presents the quantitative comparison results of semantic segmentation.
Both specialized models as well as our model have undergone training exclusively on the COCO-Stuff dataset. It is evident that our model not only surpasses specialized models in the close-set scenario, specifically the COCO-Stuff dataset, but also achieves comparable performance across other datasets that represent open-set scenarios. Notably, in the case of the VOC dataset, we observe a substantial improvement.
When compared to generalist models, our approach outperforms other competitors by a considerable margin, except in the case of Painter on the ADE-150K dataset. This is largely attributable to Painter being specifically trained on this dataset.
Interestingly, we notice that both Painter~\cite{wang2023images} and PromptDiffusion~\cite{wang2023context} lack awareness of the colors associated with unseen categories during evaluations in open-set scenarios. This is due to their reliance on example images to instruct the model regarding the color corresponding to each semantic.
In contrast, our approach establishes the color corresponding to each semantic category through text instructions, resulting in significantly superior performance. Figure~\ref{fig:refseg} illustrates several visual examples for referring segmentation to demonstrate our model's capability.





\subsection{Image Enhancement}
We evaluate the low-level vision performance of our model using the widely employed benchmarks, \ie, GoPro~\cite{Nah_2017_CVPR}, SIDD~\cite{SIDD_2018_CVPR} and CLWD~\cite{Liu_2021_WACV} dataset respectively for deblurring, denoising, and watermark removal task. The standard PSNR metric is adopted to measure the difference between the output processed image and the ground truth image. We evaluate our model's deblurring capability on the GoPro benchmark with 1280$\times$720 resolution, while for SIDD and CLWD, evaluations are conducted under 256$\times$256 resolution to align with other works. The numerical comparison is reported in Table~\ref{tab:editing-and-lowlevel}.
We have made the following observations. Firstly, specialized models trained for image editing tasks tend to exhibit poor generalization when applied to image enhancement tasks. Secondly, the generalist model Painter~\cite{wang2023images} performs better in the denoising task but encounters challenges when it comes to seamlessly integrating image editing tasks through in-context learning. Lastly, the performance of our model in image enhancement is constrained by the VAE model, which introduces information loss. We have conducted an experiment by feeding the ground truth image to the VAE model and calculating the PSNR for the output, which serves as an upper bound for our model and is indicated in parentheses.

We also present some ``in-the-wild" visual results in Figure~\ref{fig:lowlevel} to qualitatively show our model's real-world applicability in low-level vision tasks. We can observe that the resulting images have been effectively processed in line with the provided instruction, which includes sharpening, denoising, and watermark removal.

\begin{figure*}
    \centering
    \captionsetup{type=figure}
    \includegraphics[width=1\textwidth, trim={0 0 5 0},clip]{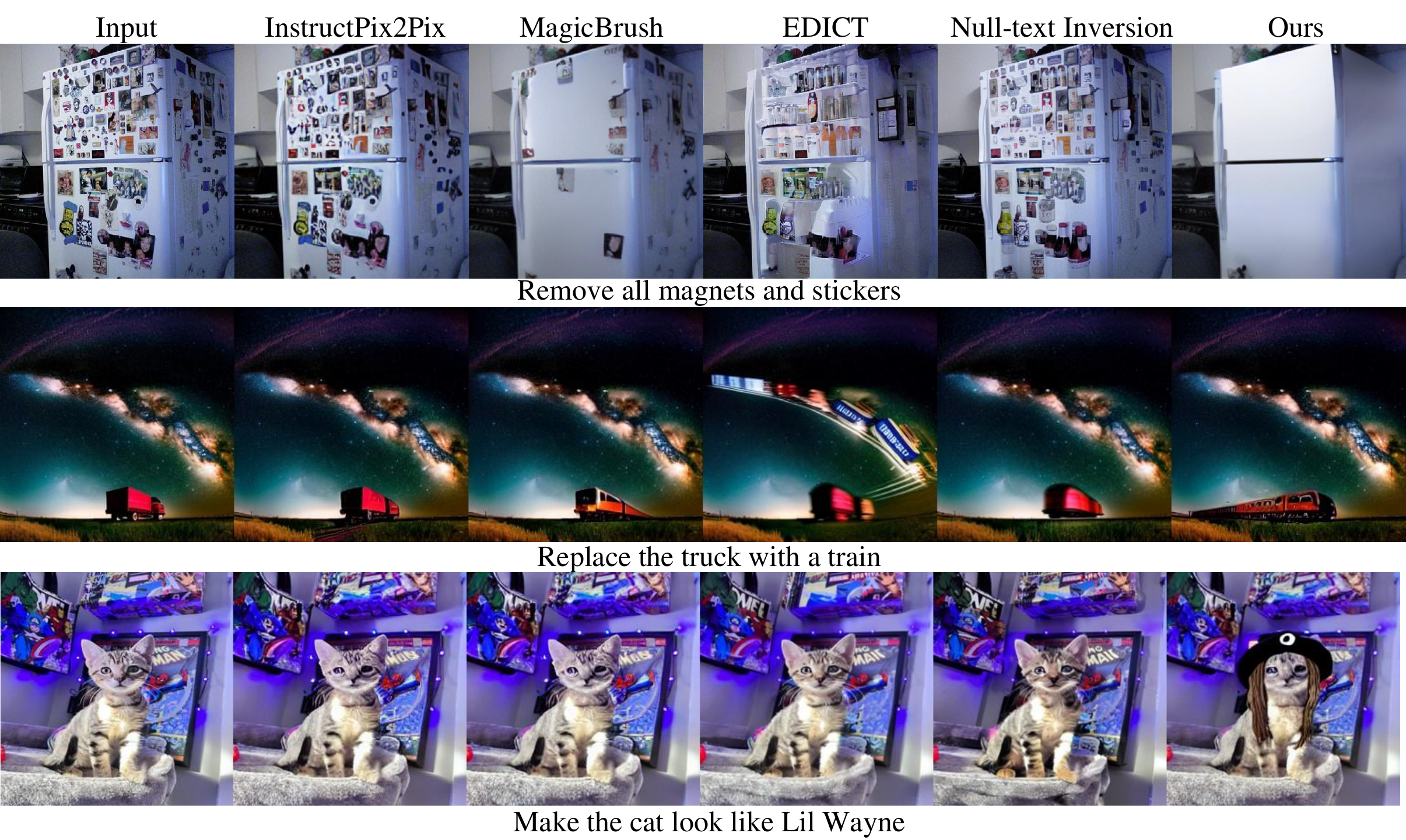}
    \captionof{figure}{Comparison between different instruction guided image editing. From left to right: input, Prompt-to-prompt~\cite{hertz2022prompt}, MagicBrush~\cite{zhang2023magicbrush}, EDICT~\cite{wallace2023edict}, Null-text Inversion~\cite{mokady2023null}, and our results.}
    \label{fig:editing-compare}
    \vspace{-1.0em}
\end{figure*}%

\begin{figure*}
    \centering
    \captionsetup{type=figure}
    \includegraphics[width=1\textwidth]{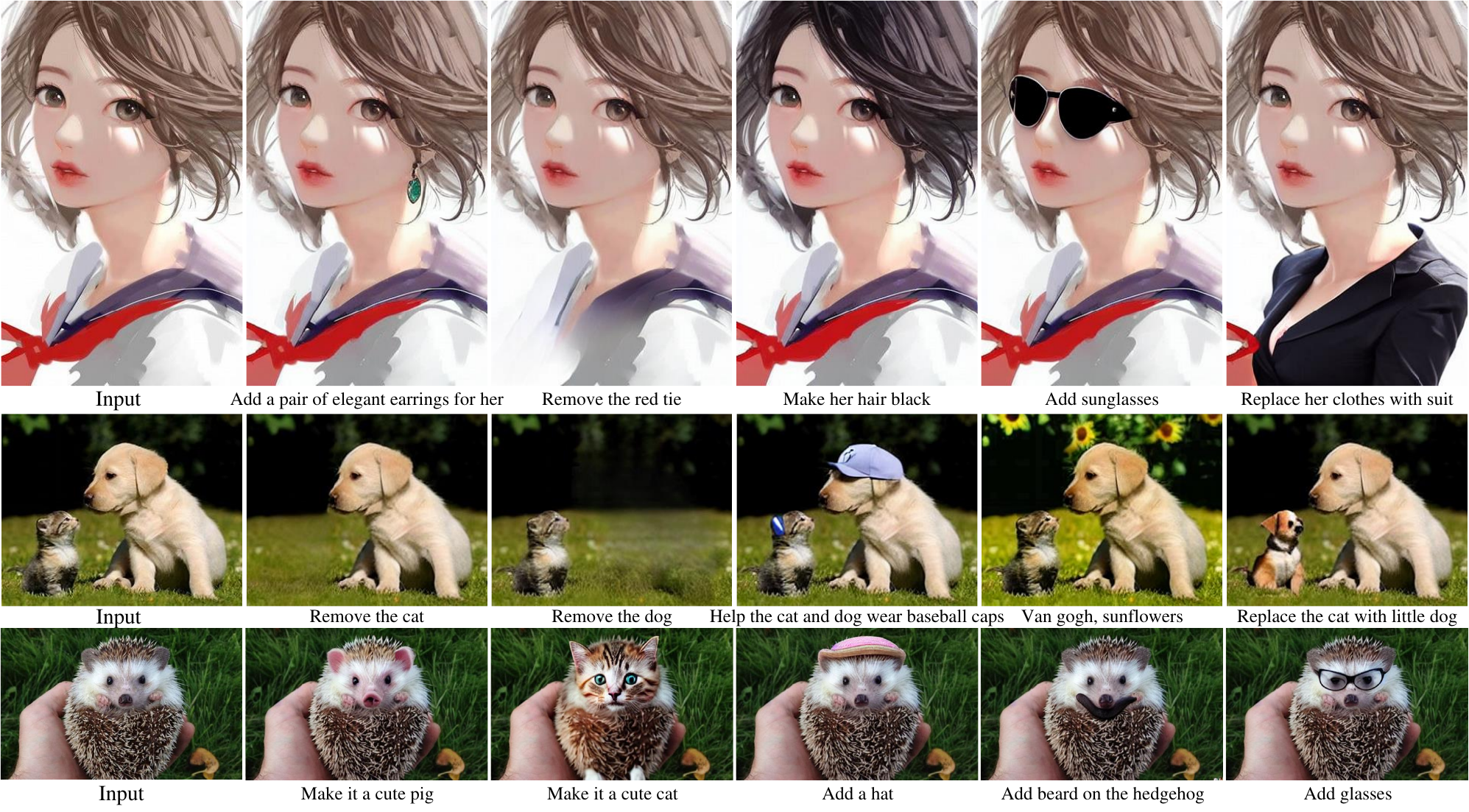}
    \captionof{figure}{Image editing results generated by our model.}
    \label{fig:editing-our}
\end{figure*}%

\subsection{Image Editing}

To better demonstrate the editing quality of our method, we build a benchmark containing 1,000 samples. Each sample contains the source image, the caption of the source image provided by BLIP2~\cite{li2023blip}, the editing instruction, and the target caption of the edited image. We manually classify each sample into one of three distinct editing scenarios: replacement, removal, and addition. This meticulous categorization aims to provide a nuanced reflection of the model's editing capabilities.
We adopt two commonly used metrics, CLIP similarity (CLIP-Sim) and Aesthetic Predictor's Score (AP)~\cite{schuhmann2022laion}, to evaluate the editing results. CLIP-Sim measures the semantic similarity between an image and a text. We utilize BLIP2~\cite{li2023blip} to obtain the caption of the input image and invoke GPT-3.5-turbo to acquire the target caption of the edited image. The CLIP-Sim score is calculated between the edited image and the target caption.
The AP score assesses the aesthetic quality of the generated images, a methodology akin to LAION-5B, which employs the CLIP+MLP Aesthetic Score Predictor. A higher quality score reflects better perceptual quality. 

We report the numerical results in Table~\ref{tab:editing-and-lowlevel}. It is important to emphasize that none of the existing generalist models have the capability to perform image editing. Compared with specific models, it is evident from the table that even with joint training, our model achieves superior CLIP-Sim compared to Instruct-Pix2Pix~\cite{brooks2023instructpix2pix} and on-par results with MagicBrush~\cite{zhang2023magicbrush}.
When assessing the editing task, it is crucial to take into account not only semantic similarity and aesthetic quality but also factors such as background consistency and manipulation accuracy. Quantitative results may be somewhat constrained for comparison. For instance, a model that fails to make substantial edits and produces an image that remains almost unchanged could potentially receive a higher score than models that have successfully carried out meaningful edits.

We further illustrate several visual examples in Figure~\ref{fig:editing-compare} compared with competitive baseline methods that have been shown impressive editing quality, including Instruct-Pix2Pix~\cite{brooks2023instructpix2pix}, MagicBrush~\cite{zhang2023magicbrush}, EDICT~\cite{wallace2023edict} and Null-text Inversion~\cite{mokady2023null}. 
It is evident that our model effectively edits the image in line with the given instructions. For instance, by following the directives, our model can successfully eliminate magnets and stickers, convert a truck into a train, and transform a cat into a particular style.
We showcase additional editing results of our model in Figure~\ref{fig:editing-our}, further highlighting the remarkably precise editing quality achieved.
Given a source image, our model is capable of successfully adding, removing, and replacing elements. The image undergoes precise editing based on the prompt while maintaining the integrity of the background and preserving intricate details.

\renewcommand{\arraystretch}{1.2}
 \begin{table*}[]
\setlength\tabcolsep{5.3pt}
\centering
\caption{Ablation study on the instruction. The term "Simple Instruction" denotes coarse-grained instructions like "semantic segmentation" or "keypoint detection". In contrast, our approach utilizes highly detailed and more flexible instructions. The training datasets include COCO for keypoint detection and COCO-Stuff for semantic segmentation.}
\vspace{-0.7em}
\begin{tabular}{c||c|cc||c|cccccc}
\bottomrule
\multirow{2}{*}{Method} & \multicolumn{3}{c||}{Keypoint Detection} & \multicolumn{6}{c}{Semantic Segmentation} \\
\cline{2-10}
& COCO & HumanArt & AP-10K & COCO-Stuff & ADE-847 & PC-459 & ADE-150 & PC-59 & VOC \\ \hline
Simple Instruction & 22.7 & 7.0 & 5.2 & 41.28 & 1.39 & 3.96 & 13.65 & 18.49 & 20.22 \\
Ours & 71.2 & 51.4 & 15.9 & 53.17 & 19.68 & 28.29 & 33.62 & 59.00 & 72.55\\
\toprule
\end{tabular}
\label{tab:detailed-instruction}
\vspace{-1.15em}
\end{table*}


\subsection{The Benefit of Highly Detailed Instruction}
Our hypothesis is that the ability to generalize is the skill of learning through understanding the specific meanings of individual elements rather than memorizing entire instructions.
Unlike previous unified models like Pix2seq~\cite{chen2021pix2seq} and Unified-IO~\cite{lu2022unified}, which simply treat natural language as task indicators, our approach employs detailed descriptions for each task as instructions.
 Such detailed instructions enable the model to understand comprehensively and then prioritize accurate execution instead of simple instructions that favor mimicking. 
To show this, we try to replace our instructions within our framework with simpler task indicators, such as "semantic segmentation" and "keypoint detection," while assigning fixed colors to each keypoint or object class. As demonstrated in Table~\ref{tab:detailed-instruction}, the results of simple instructions are extremely poor, especially when handling new types of keypoints or novel object categories. This highlights that our detailed instructions provide enhanced flexibility and adaptability in the open domain.

\subsection{The Benefit of Multi-task Training}
\begin{figure}
    \centering
    \captionsetup{type=figure}
    \includegraphics[width=0.48\textwidth]{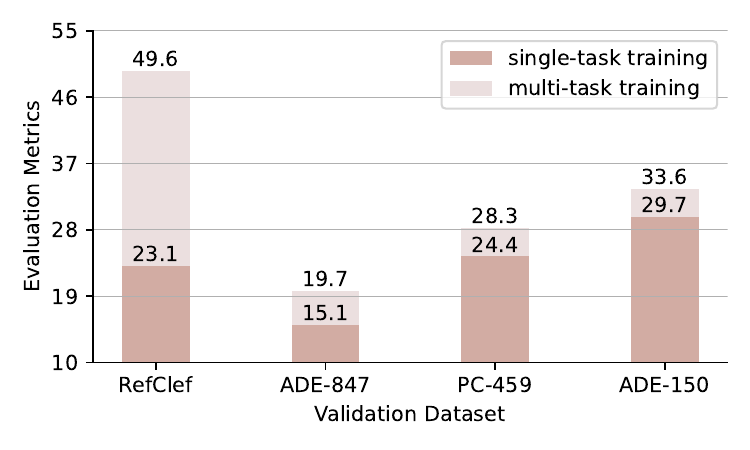}
    \vspace{-2.0em}
    \captionof{figure}{Ablation study on multi-task training. We evaluate our models on four unseen datasets, RefClef, ADE-847, PC-459, and ADE-150. It demonstrates that joint training significantly enhances the capability to handle open-set scenarios.}
    \label{fig:multi-task}
    \vspace{-1.7em}
\end{figure}

\begin{figure}
    \centering
    \captionsetup{type=figure}
    \includegraphics[width=0.477\textwidth]{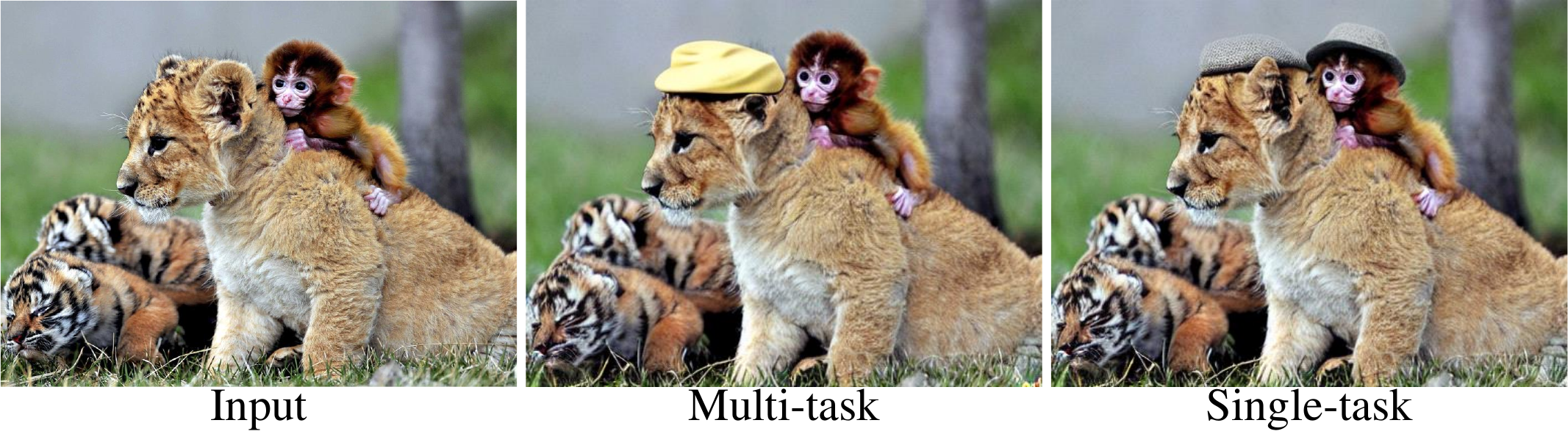}
    \captionof{figure}{Multi-task training's effect on editing. When given the instruction ``put a hat on the leopard'', joint training with other tasks can lead to more precise editing outcomes in comparison to single-task training. }
    \label{fig:multi-task-edit}
\end{figure}

Multi-task learning has grown increasingly popular, enabling models to achieve greater generalization by concurrently addressing multiple related tasks during training. This approach often results in improved model generalization performance compared to specialized single-task training. To further provide empirical validation for this observation, we experiment with our model when trained only on the segmentation dataset and report the performance difference in Figure~\ref{fig:multi-task}. This illustration compares the results of our single-task model and multi-task model evaluated over four unseen test datasets. 
It is evident that our jointly trained model performs significantly better in open-domain testing scenarios compared to the specific models. 
Furthermore, we observe that this benefit also extends to image editing. In Figure~\ref{fig:multi-task-edit}, the visual comparison demonstrates that when integrated with other tasks, the model can more effectively discern which objects require editing, potentially benefiting from the integration of referring segmentation. 

\begin{figure}
    \centering
    \captionsetup{type=figure}
    \includegraphics[width=0.45\textwidth]{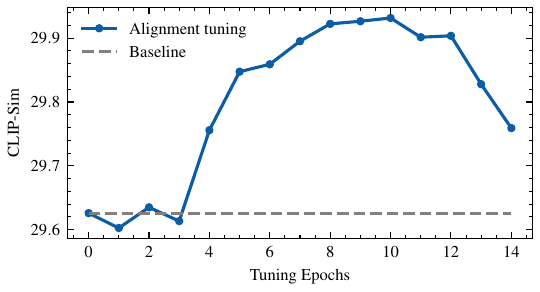}
    \captionof{figure}{Effect of human alignment. Further fine-tuning on selected human alignment data enhances the CLIP-Sim metric, reaching its peak after approximately 10 epochs.}
    \label{fig:instruction}
    \vspace{-0.2em}
\end{figure}%

\subsection{The Benefit of Human Alignment}

Our model undergoes a subsequent fine-tuning phase using a filtered dataset with human alignment. In this evaluation, we examine its effectiveness and present the fine-tuning progress in Figure~\ref{fig:instruction}, which showcases the relationship between CLIP-Sim performance and the number of epochs.
Initially, the CLIP-Sim score stands at 29.6. Remarkably, we observe a noticeable enhancement in image-text alignment, which increases from 29.6 to 29.9 over approximately 10 epochs. It is important to highlight the significance of this improvement, particularly considering that the dataset consists of only 1000 samples.

\subsection{Generalization Capability to Unseen Tasks}
\begin{figure}
    \centering
    \captionsetup{type=figure}
    \includegraphics[width=0.478\textwidth]{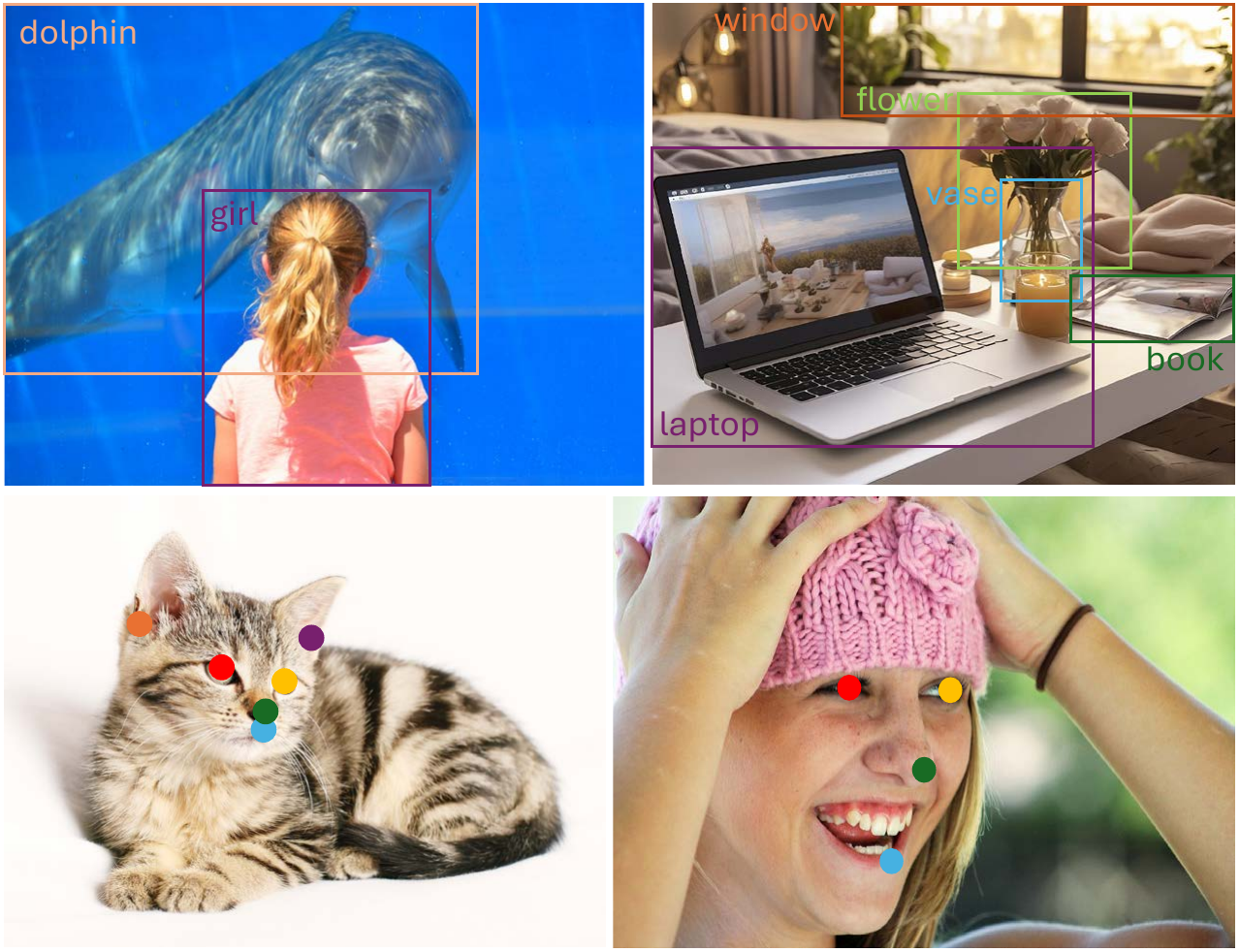}
    \captionof{figure}{Adopt InstructDiffusion to unseen tasks, including detection, classification, and face alignment.}
    \label{fig:unseen}
    \vspace{-0.6em}
\end{figure}

We demonstrate that our model exhibits a degree of Artificial General Intelligence (AGI) capabilities by leveraging the wealth of tasks and diverse datasets through this highly detailed instruction-following format.
We validate its capacity to handle tasks that were not part of its training repertoire, including image detection, classification, and even intricate fine-grained tasks like face alignment in Figure~\ref{fig:unseen}. 
In the context of detection and classification, we employ a prompt that resembles referring segmentation, enabling us to derive the bounding box coordinates by identifying the top, bottom, left, and right boundaries of the marked region. Moreover, we can also verify the class label using a versatile prompt structure.
In the realm of face alignment, our approach involves directly instructing our model to encircle the specific facial region of interest, such as the nose or right ear. Remarkably, we have found that this approach performs admirably even when applied to animal faces.
We argue that this underscores its versatility in adapting to new challenges beyond its initial training scope.

\section{Discussion and conclusion}
\label{sec:conclusion}
In conclusion, this paper presents InstructDiffusion, a novel and unifying framework for aligning computer vision tasks with human instructions. InstructDiffusion treats all computer vision tasks as image generation, with a focus on three types of output formats: 3-channel RGB images, binary masks, and keypoints. We demonstrated that our approach achieves good performance in individual tasks, and joint training of multiple tasks enhances the generalization ability. Remarkably, InstructDiffusion exhibits AGI capabilities to some extent, handling tasks not seen during training and outperforming previous methods on unseen datasets. This research marks a significant step towards a generalist modeling interface for vision tasks and sets the stage for future advancements in the pursuit of artificial general intelligence in computer vision.

In future work, we plan to focus on the following aspects to further improve the performance and capabilities of InstructDiffusion: 1)
Improve the unified representation: We aim to explore alternative encoding schemes and techniques to better represent a more diverse range of outputs associated with various computer vision tasks.  2) Investigate the role of self-supervised and unsupervised learning: To enhance the generalization ability of InstructDiffusion, we will explore the use of self-supervised and unsupervised learning techniques to leverage large-scale unlabeled data for model training and adaptation.

{\small
\bibliographystyle{ieee_fullname}
\bibliography{egbib}

\begin{thebibliography}{10}\itemsep=-1pt

\bibitem{SIDD_2018_CVPR}
Abdelrahman Abdelhamed, Stephen Lin, and Michael~S. Brown.
\newblock A high-quality denoising dataset for smartphone cameras.
\newblock In {\em IEEE Conference on Computer Vision and Pattern Recognition
  (CVPR)}, June 2018.

\bibitem{alayrac2022flamingo}
Jean-Baptiste Alayrac, Jeff Donahue, Pauline Luc, Antoine Miech, Iain Barr,
  Yana Hasson, Karel Lenc, Arthur Mensch, Katherine Millican, Malcolm Reynolds,
  et~al.
\newblock Flamingo: a visual language model for few-shot learning.
\newblock {\em Advances in Neural Information Processing Systems},
  35:23716--23736, 2022.

\bibitem{Andriluka14mpii}
Mykhaylo Andriluka, Leonid Pishchulin, Peter Gehler, and Bernt Schiele.
\newblock 2d human pose estimation: New benchmark and state of the art
  analysis.
\newblock In {\em Proceedings of the IEEE Conference on computer Vision and
  Pattern Recognition}, pages 3686--3693, 2014.

\bibitem{anil2023palm}
Rohan Anil, Andrew~M Dai, Orhan Firat, Melvin Johnson, Dmitry Lepikhin,
  Alexandre Passos, Siamak Shakeri, Emanuel Taropa, Paige Bailey, Zhifeng Chen,
  et~al.
\newblock Palm 2 technical report.
\newblock {\em arXiv preprint arXiv:2305.10403}, 2023.

\bibitem{bar2022visual}
Amir Bar, Yossi Gandelsman, Trevor Darrell, Amir Globerson, and Alexei Efros.
\newblock Visual prompting via image inpainting.
\newblock {\em Advances in Neural Information Processing Systems},
  35:25005--25017, 2022.

\bibitem{brooks2023instructpix2pix}
Tim Brooks, Aleksander Holynski, and Alexei~A Efros.
\newblock Instructpix2pix: Learning to follow image editing instructions.
\newblock In {\em Proceedings of the IEEE/CVF Conference on Computer Vision and
  Pattern Recognition}, pages 18392--18402, 2023.

\bibitem{brown2020language}
Tom Brown, Benjamin Mann, Nick Ryder, Melanie Subbiah, Jared~D Kaplan, Prafulla
  Dhariwal, Arvind Neelakantan, Pranav Shyam, Girish Sastry, Amanda Askell,
  et~al.
\newblock Language models are few-shot learners.
\newblock {\em Advances in neural information processing systems},
  33:1877--1901, 2020.

\bibitem{caesar2018cvpr}
Holger Caesar, Jasper Uijlings, and Vittorio Ferrari.
\newblock Coco-stuff: Thing and stuff classes in context.
\newblock In {\em Computer vision and pattern recognition (CVPR), 2018 IEEE
  conference on}. IEEE, 2018.

\bibitem{chen2022simple}
Liangyu Chen, Xiaojie Chu, Xiangyu Zhang, and Jian Sun.
\newblock Simple baselines for image restoration.
\newblock {\em arXiv preprint arXiv:2204.04676}, 2022.

\bibitem{chen2021pix2seq}
Ting Chen, Saurabh Saxena, Lala Li, David~J Fleet, and Geoffrey Hinton.
\newblock Pix2seq: A language modeling framework for object detection.
\newblock {\em arXiv preprint arXiv:2109.10852}, 2021.

\bibitem{chen2022unified}
Ting Chen, Saurabh Saxena, Lala Li, Tsung-Yi Lin, David~J Fleet, and Geoffrey~E
  Hinton.
\newblock A unified sequence interface for vision tasks.
\newblock {\em Advances in Neural Information Processing Systems},
  35:31333--31346, 2022.

\bibitem{chowdhery2022palm}
Aakanksha Chowdhery, Sharan Narang, Jacob Devlin, Maarten Bosma, Gaurav Mishra,
  Adam Roberts, Paul Barham, Hyung~Won Chung, Charles Sutton, Sebastian
  Gehrmann, et~al.
\newblock Palm: Scaling language modeling with pathways.
\newblock {\em arXiv preprint arXiv:2204.02311}, 2022.

\bibitem{devlin2018bert}
Jacob Devlin, Ming-Wei Chang, Kenton Lee, and Kristina Toutanova.
\newblock Bert: Pre-training of deep bidirectional transformers for language
  understanding.
\newblock {\em arXiv preprint arXiv:1810.04805}, 2018.

\bibitem{everingham2012pascal}
Mark Everingham and John Winn.
\newblock The pascal visual object classes challenge 2012 (voc2012) development
  kit.
\newblock {\em Pattern Anal. Stat. Model. Comput. Learn., Tech. Rep},
  2007(1-45):5, 2012.

\bibitem{frome2013devise}
Andrea Frome, Greg~S Corrado, Jon Shlens, Samy Bengio, Jeff Dean, Marc'Aurelio
  Ranzato, and Tomas Mikolov.
\newblock Devise: A deep visual-semantic embedding model.
\newblock {\em Advances in neural information processing systems}, 26, 2013.

\bibitem{Geng23PCT}
Zigang Geng, Chunyu Wang, Yixuan Wei, Ze Liu, Houqiang Li, and Han Hu.
\newblock Human pose as compositional tokens.
\newblock In {\em {CVPR}}, 2023.

\bibitem{goodfellow2020generative}
Ian Goodfellow, Jean Pouget-Abadie, Mehdi Mirza, Bing Xu, David Warde-Farley,
  Sherjil Ozair, Aaron Courville, and Yoshua Bengio.
\newblock Generative adversarial networks.
\newblock {\em Communications of the ACM}, 63(11):139--144, 2020.

\bibitem{gu2020giqa}
Shuyang Gu, Jianmin Bao, Dong Chen, and Fang Wen.
\newblock Giqa: Generated image quality assessment.
\newblock In {\em European conference on computer vision}, pages 369--385.
  Springer, 2020.

\bibitem{gu2022vector}
Shuyang Gu, Dong Chen, Jianmin Bao, Fang Wen, Bo Zhang, Dongdong Chen, Lu Yuan,
  and Baining Guo.
\newblock Vector quantized diffusion model for text-to-image synthesis.
\newblock In {\em Proceedings of the IEEE/CVF Conference on Computer Vision and
  Pattern Recognition}, pages 10696--10706, 2022.

\bibitem{gupta2022towards}
Tanmay Gupta, Amita Kamath, Aniruddha Kembhavi, and Derek Hoiem.
\newblock Towards general purpose vision systems: An end-to-end task-agnostic
  vision-language architecture.
\newblock In {\em Proceedings of the IEEE/CVF Conference on Computer Vision and
  Pattern Recognition}, pages 16399--16409, 2022.

\bibitem{hertz2022prompt}
Amir Hertz, Ron Mokady, Jay Tenenbaum, Kfir Aberman, Yael Pritch, and Daniel
  Cohen-or.
\newblock Prompt-to-prompt image editing with cross-attention control.
\newblock In {\em The Eleventh International Conference on Learning
  Representations}, 2022.

\bibitem{ho2020denoising}
Jonathan Ho, Ajay Jain, and Pieter Abbeel.
\newblock Denoising diffusion probabilistic models.
\newblock {\em Advances in Neural Information Processing Systems},
  33:6840--6851, 2020.

\bibitem{ho2022classifier}
Jonathan Ho and Tim Salimans.
\newblock Classifier-free diffusion guidance.
\newblock {\em arXiv preprint arXiv:2207.12598}, 2022.

\bibitem{huang2023language}
Shaohan Huang, Li Dong, Wenhui Wang, Yaru Hao, Saksham Singhal, Shuming Ma,
  Tengchao Lv, Lei Cui, Owais~Khan Mohammed, Qiang Liu, et~al.
\newblock Language is not all you need: Aligning perception with language
  models.
\newblock {\em arXiv preprint arXiv:2302.14045}, 2023.

\bibitem{jaegle2021perceiver}
Andrew Jaegle, Felix Gimeno, Andy Brock, Oriol Vinyals, Andrew Zisserman, and
  Joao Carreira.
\newblock Perceiver: General perception with iterative attention.
\newblock In {\em International conference on machine learning}, pages
  4651--4664. PMLR, 2021.

\bibitem{jia2021scaling}
Chao Jia, Yinfei Yang, Ye Xia, Yi-Ting Chen, Zarana Parekh, Hieu Pham, Quoc Le,
  Yun-Hsuan Sung, Zhen Li, and Tom Duerig.
\newblock Scaling up visual and vision-language representation learning with
  noisy text supervision.
\newblock In {\em International conference on machine learning}, pages
  4904--4916. PMLR, 2021.

\bibitem{ju2023humanart}
Xuan Ju, Ailing Zeng, Wang Jianan, Xu Qiang, and Zhang Lei.
\newblock Human-art: A versatile human-centric dataset bridging natural and
  artificial scenes.
\newblock In {\em Proceedings of the IEEE/CVF Conference on Computer Vision and
  Pattern Recognition (CVPR)}, 2023.

\bibitem{karras2019style}
Tero Karras, Samuli Laine, and Timo Aila.
\newblock A style-based generator architecture for generative adversarial
  networks.
\newblock In {\em Proceedings of the IEEE/CVF conference on computer vision and
  pattern recognition}, pages 4401--4410, 2019.

\bibitem{karras2020analyzing}
Tero Karras, Samuli Laine, Miika Aittala, Janne Hellsten, Jaakko Lehtinen, and
  Timo Aila.
\newblock Analyzing and improving the image quality of stylegan.
\newblock In {\em Proceedings of the IEEE/CVF conference on computer vision and
  pattern recognition}, pages 8110--8119, 2020.

\bibitem{kazemzadeh2014referitgame}
Sahar Kazemzadeh, Vicente Ordonez, Mark Matten, and Tamara Berg.
\newblock Referitgame: Referring to objects in photographs of natural scenes.
\newblock In {\em Proceedings of the 2014 conference on empirical methods in
  natural language processing (EMNLP)}, pages 787--798, 2014.

\bibitem{kirillov2023segment}
Alexander Kirillov, Eric Mintun, Nikhila Ravi, Hanzi Mao, Chloe Rolland, Laura
  Gustafson, Tete Xiao, Spencer Whitehead, Alexander~C Berg, Wan-Yen Lo, et~al.
\newblock Segment anything.
\newblock {\em arXiv preprint arXiv:2304.02643}, 2023.

\bibitem{kuznetsova2020open}
Alina Kuznetsova, Hassan Rom, Neil Alldrin, Jasper Uijlings, Ivan Krasin, Jordi
  Pont-Tuset, Shahab Kamali, Stefan Popov, Matteo Malloci, Alexander
  Kolesnikov, et~al.
\newblock The open images dataset v4.
\newblock {\em International Journal of Computer Vision}, 128(7):1956--1981,
  2020.

\bibitem{li2023uni}
Hao Li, Jinguo Zhu, Xiaohu Jiang, Xizhou Zhu, Hongsheng Li, Chun Yuan, Xiaohua
  Wang, Yu Qiao, Xiaogang Wang, Wenhai Wang, et~al.
\newblock Uni-perceiver v2: A generalist model for large-scale vision and
  vision-language tasks.
\newblock In {\em Proceedings of the IEEE/CVF Conference on Computer Vision and
  Pattern Recognition}, pages 2691--2700, 2023.

\bibitem{li2023blip}
Junnan Li, Dongxu Li, Silvio Savarese, and Steven Hoi.
\newblock Blip-2: Bootstrapping language-image pre-training with frozen image
  encoders and large language models.
\newblock {\em arXiv preprint arXiv:2301.12597}, 2023.

\bibitem{li2022blip}
Junnan Li, Dongxu Li, Caiming Xiong, and Steven Hoi.
\newblock Blip: Bootstrapping language-image pre-training for unified
  vision-language understanding and generation.
\newblock In {\em International Conference on Machine Learning}, pages
  12888--12900. PMLR, 2022.

\bibitem{li2021align}
Junnan Li, Ramprasaath Selvaraju, Akhilesh Gotmare, Shafiq Joty, Caiming Xiong,
  and Steven Chu~Hong Hoi.
\newblock Align before fuse: Vision and language representation learning with
  momentum distillation.
\newblock {\em Advances in neural information processing systems},
  34:9694--9705, 2021.

\bibitem{li2018crowdpose}
Jiefeng Li, Can Wang, Hao Zhu, Yihuan Mao, Hao-Shu Fang, and Cewu Lu.
\newblock Crowdpose: Efficient crowded scenes pose estimation and a new
  benchmark.
\newblock In {\em {CVPR}}, 2019.

\bibitem{li2022grounded}
Liunian~Harold Li, Pengchuan Zhang, Haotian Zhang, Jianwei Yang, Chunyuan Li,
  Yiwu Zhong, Lijuan Wang, Lu Yuan, Lei Zhang, Jenq-Neng Hwang, et~al.
\newblock Grounded language-image pre-training.
\newblock In {\em Proceedings of the IEEE/CVF Conference on Computer Vision and
  Pattern Recognition}, pages 10965--10975, 2022.

\bibitem{li2020oscar}
Xiujun Li, Xi Yin, Chunyuan Li, Pengchuan Zhang, Xiaowei Hu, Lei Zhang, Lijuan
  Wang, Houdong Hu, Li Dong, Furu Wei, et~al.
\newblock Oscar: Object-semantics aligned pre-training for vision-language
  tasks.
\newblock In {\em Computer Vision--ECCV 2020: 16th European Conference,
  Glasgow, UK, August 23--28, 2020, Proceedings, Part XXX 16}, pages 121--137.
  Springer, 2020.

\bibitem{liang2023open}
Feng Liang, Bichen Wu, Xiaoliang Dai, Kunpeng Li, Yinan Zhao, Hang Zhang,
  Peizhao Zhang, Peter Vajda, and Diana Marculescu.
\newblock Open-vocabulary semantic segmentation with mask-adapted clip.
\newblock In {\em Proceedings of the IEEE/CVF Conference on Computer Vision and
  Pattern Recognition}, pages 7061--7070, 2023.

\bibitem{lin2014microsoft}
Tsung-Yi Lin, Michael Maire, Serge Belongie, James Hays, Pietro Perona, Deva
  Ramanan, Piotr Doll{\'a}r, and C~Lawrence Zitnick.
\newblock Microsoft coco: Common objects in context.
\newblock In {\em European conference on computer vision}, pages 740--755.
  Springer, 2014.

\bibitem{GRES}
Chang Liu, Henghui Ding, and Xudong Jiang.
\newblock {GRES}: Generalized referring expression segmentation.
\newblock In {\em CVPR}, 2023.

\bibitem{liu2023visual}
Haotian Liu, Chunyuan Li, Qingyang Wu, and Yong~Jae Lee.
\newblock Visual instruction tuning.
\newblock {\em arXiv preprint arXiv:2304.08485}, 2023.

\bibitem{Liu_2021_WACV}
Yang Liu, Zhen Zhu, and Xiang Bai.
\newblock Wdnet: Watermark-decomposition network for visible watermark removal.
\newblock In {\em 2021 {IEEE/CVF} Winter Conference on Applications of Computer
  Vision (WACV)}. {IEEE}, 2021.

\bibitem{lu2019vilbert}
Jiasen Lu, Dhruv Batra, Devi Parikh, and Stefan Lee.
\newblock Vilbert: Pretraining task-agnostic visiolinguistic representations
  for vision-and-language tasks.
\newblock {\em Advances in neural information processing systems}, 32, 2019.

\bibitem{lu2022unified}
Jiasen Lu, Christopher Clark, Rowan Zellers, Roozbeh Mottaghi, and Aniruddha
  Kembhavi.
\newblock Unified-io: A unified model for vision, language, and multi-modal
  tasks.
\newblock {\em arXiv preprint arXiv:2206.08916}, 2022.

\bibitem{mao2016generation}
Junhua Mao, Jonathan Huang, Alexander Toshev, Oana Camburu, Alan~L Yuille, and
  Kevin Murphy.
\newblock Generation and comprehension of unambiguous object descriptions.
\newblock In {\em Proceedings of the IEEE conference on computer vision and
  pattern recognition}, pages 11--20, 2016.

\bibitem{mokady2023null}
Ron Mokady, Amir Hertz, Kfir Aberman, Yael Pritch, and Daniel Cohen-Or.
\newblock Null-text inversion for editing real images using guided diffusion
  models.
\newblock In {\em Proceedings of the IEEE/CVF Conference on Computer Vision and
  Pattern Recognition}, pages 6038--6047, 2023.

\bibitem{mottaghi2014role}
Roozbeh Mottaghi, Xianjie Chen, Xiaobai Liu, Nam-Gyu Cho, Seong-Whan Lee, Sanja
  Fidler, Raquel Urtasun, and Alan Yuille.
\newblock The role of context for object detection and semantic segmentation in
  the wild.
\newblock In {\em Proceedings of the IEEE conference on computer vision and
  pattern recognition}, pages 891--898, 2014.

\bibitem{Nah_2019_CVPR_Workshops_REDS}
Seungjun Nah, Sungyong Baik, Seokil Hong, Gyeongsik Moon, Sanghyun Son, Radu
  Timofte, and Kyoung~Mu Lee.
\newblock Ntire 2019 challenge on video deblurring and super-resolution:
  Dataset and study.
\newblock In {\em CVPR Workshops}, June 2019.

\bibitem{Nah_2017_CVPR}
Seungjun Nah, Tae~Hyun Kim, and Kyoung~Mu Lee.
\newblock Deep multi-scale convolutional neural network for dynamic scene
  deblurring.
\newblock In {\em CVPR}, July 2017.

\bibitem{openai2023gpt4}
OpenAI.
\newblock Gpt-4 technical report, 2023.

\bibitem{ouyang2022training}
Long Ouyang, Jeffrey Wu, Xu Jiang, Diogo Almeida, Carroll Wainwright, Pamela
  Mishkin, Chong Zhang, Sandhini Agarwal, Katarina Slama, Alex Ray, et~al.
\newblock Training language models to follow instructions with human feedback.
\newblock {\em Advances in Neural Information Processing Systems},
  35:27730--27744, 2022.

\bibitem{peng2023kosmos}
Zhiliang Peng, Wenhui Wang, Li Dong, Yaru Hao, Shaohan Huang, Shuming Ma, and
  Furu Wei.
\newblock Kosmos-2: Grounding multimodal large language models to the world.
\newblock {\em arXiv preprint arXiv:2306.14824}, 2023.

\bibitem{radford2021learning}
Alec Radford, Jong~Wook Kim, Chris Hallacy, Aditya Ramesh, Gabriel Goh,
  Sandhini Agarwal, Girish Sastry, Amanda Askell, Pamela Mishkin, Jack Clark,
  et~al.
\newblock Learning transferable visual models from natural language
  supervision.
\newblock In {\em International Conference on Machine Learning}, pages
  8748--8763. PMLR, 2021.

\bibitem{radford2018improving}
Alec Radford, Karthik Narasimhan, Tim Salimans, Ilya Sutskever, et~al.
\newblock Improving language understanding by generative pre-training.
\newblock 2018.

\bibitem{radford2019language}
Alec Radford, Jeffrey Wu, Rewon Child, David Luan, Dario Amodei, Ilya
  Sutskever, et~al.
\newblock Language models are unsupervised multitask learners.
\newblock {\em OpenAI blog}, 1(8):9, 2019.

\bibitem{razavi2019generating}
Ali Razavi, Aaron Van~den Oord, and Oriol Vinyals.
\newblock Generating diverse high-fidelity images with vq-vae-2.
\newblock {\em Advances in neural information processing systems}, 32, 2019.

\bibitem{reed2022generalist}
Scott Reed, Konrad Zolna, Emilio Parisotto, Sergio~Gomez Colmenarejo, Alexander
  Novikov, Gabriel Barth-Maron, Mai Gimenez, Yury Sulsky, Jackie Kay,
  Jost~Tobias Springenberg, et~al.
\newblock A generalist agent.
\newblock {\em arXiv preprint arXiv:2205.06175}, 2022.

\bibitem{rombach2022high}
Robin Rombach, Andreas Blattmann, Dominik Lorenz, Patrick Esser, and Bj{\"o}rn
  Ommer.
\newblock High-resolution image synthesis with latent diffusion models.
\newblock In {\em Proceedings of the IEEE/CVF Conference on Computer Vision and
  Pattern Recognition}, pages 10684--10695, 2022.

\bibitem{schuhmann2022laion}
Christoph Schuhmann, Romain Beaumont, Richard Vencu, Cade Gordon, Ross
  Wightman, Mehdi Cherti, Theo Coombes, Aarush Katta, Clayton Mullis, Mitchell
  Wortsman, et~al.
\newblock Laion-5b: An open large-scale dataset for training next generation
  image-text models.
\newblock {\em Advances in Neural Information Processing Systems},
  35:25278--25294, 2022.

\bibitem{shao2021intern}
Jing Shao, Siyu Chen, Yangguang Li, Kun Wang, Zhenfei Yin, Yinan He, Jianing
  Teng, Qinghong Sun, Mengya Gao, Jihao Liu, et~al.
\newblock Intern: A new learning paradigm towards general vision.
\newblock {\em arXiv preprint arXiv:2111.08687}, 2021.

\bibitem{shi2020gierbenchmark}
Jing Shi, Ning Xu, Trung Bui, Franck Dernoncourt, Zheng Wen, and Chenliang Xu.
\newblock A benchmark and baseline for language-driven image editing.
\newblock In {\em Proceedings of the Asian Conference on Computer Vision},
  2020.

\bibitem{singh2022flava}
Amanpreet Singh, Ronghang Hu, Vedanuj Goswami, Guillaume Couairon, Wojciech
  Galuba, Marcus Rohrbach, and Douwe Kiela.
\newblock Flava: A foundational language and vision alignment model.
\newblock In {\em Proceedings of the IEEE/CVF Conference on Computer Vision and
  Pattern Recognition}, pages 15638--15650, 2022.

\bibitem{song2020score}
Yang Song, Jascha Sohl-Dickstein, Diederik~P Kingma, Abhishek Kumar, Stefano
  Ermon, and Ben Poole.
\newblock Score-based generative modeling through stochastic differential
  equations.
\newblock {\em arXiv preprint arXiv:2011.13456}, 2020.

\bibitem{su2019vl}
Weijie Su, Xizhou Zhu, Yue Cao, Bin Li, Lewei Lu, Furu Wei, and Jifeng Dai.
\newblock Vl-bert: Pre-training of generic visual-linguistic representations.
\newblock {\em arXiv preprint arXiv:1908.08530}, 2019.

\bibitem{suvorov2022resolution}
Roman Suvorov, Elizaveta Logacheva, Anton Mashikhin, Anastasia Remizova,
  Arsenii Ashukha, Aleksei Silvestrov, Naejin Kong, Harshith Goka, Kiwoong
  Park, and Victor Lempitsky.
\newblock Resolution-robust large mask inpainting with fourier convolutions.
\newblock In {\em Proceedings of the IEEE/CVF Winter Conference on Applications
  of Computer Vision}, pages 2149--2159, 2022.

\bibitem{touvron2023llama}
Hugo Touvron, Thibaut Lavril, Gautier Izacard, Xavier Martinet, Marie-Anne
  Lachaux, Timoth{\'e}e Lacroix, Baptiste Rozi{\`e}re, Naman Goyal, Eric
  Hambro, Faisal Azhar, et~al.
\newblock Llama: Open and efficient foundation language models.
\newblock {\em arXiv preprint arXiv:2302.13971}, 2023.

\bibitem{tsimpoukelli2021multimodal}
Maria Tsimpoukelli, Jacob~L Menick, Serkan Cabi, SM Eslami, Oriol Vinyals, and
  Felix Hill.
\newblock Multimodal few-shot learning with frozen language models.
\newblock {\em Advances in Neural Information Processing Systems}, 34:200--212,
  2021.

\bibitem{van2017neural}
Aaron Van Den~Oord, Oriol Vinyals, et~al.
\newblock Neural discrete representation learning.
\newblock {\em Advances in neural information processing systems}, 30, 2017.

\bibitem{vaswani2017attention}
Ashish Vaswani, Noam Shazeer, Niki Parmar, Jakob Uszkoreit, Llion Jones,
  Aidan~N Gomez, {\L}ukasz Kaiser, and Illia Polosukhin.
\newblock Attention is all you need.
\newblock {\em Advances in neural information processing systems}, 30, 2017.

\bibitem{wallace2023edict}
Bram Wallace, Akash Gokul, and Nikhil Naik.
\newblock Edict: Exact diffusion inversion via coupled transformations.
\newblock In {\em Proceedings of the IEEE/CVF Conference on Computer Vision and
  Pattern Recognition}, pages 22532--22541, 2023.

\bibitem{wang2022omnivl}
Junke Wang, Dongdong Chen, Zuxuan Wu, Chong Luo, Luowei Zhou, Yucheng Zhao,
  Yujia Xie, Ce Liu, Yu-Gang Jiang, and Lu Yuan.
\newblock Omnivl: One foundation model for image-language and video-language
  tasks.
\newblock {\em Advances in neural information processing systems},
  35:5696--5710, 2022.

\bibitem{wang2022git}
Jianfeng Wang, Zhengyuan Yang, Xiaowei Hu, Linjie Li, Kevin Lin, Zhe Gan,
  Zicheng Liu, Ce Liu, and Lijuan Wang.
\newblock Git: A generative image-to-text transformer for vision and language.
\newblock {\em arXiv preprint arXiv:2205.14100}, 2022.

\bibitem{wang2016learning}
Liwei Wang, Yin Li, and Svetlana Lazebnik.
\newblock Learning deep structure-preserving image-text embeddings.
\newblock In {\em Proceedings of the IEEE conference on computer vision and
  pattern recognition}, pages 5005--5013, 2016.

\bibitem{wang2022ofa}
Peng Wang, An Yang, Rui Men, Junyang Lin, Shuai Bai, Zhikang Li, Jianxin Ma,
  Chang Zhou, Jingren Zhou, and Hongxia Yang.
\newblock Ofa: Unifying architectures, tasks, and modalities through a simple
  sequence-to-sequence learning framework.
\newblock In {\em International Conference on Machine Learning}, pages
  23318--23340. PMLR, 2022.

\bibitem{wang2022image}
Wenhui Wang, Hangbo Bao, Li Dong, Johan Bjorck, Zhiliang Peng, Qiang Liu, Kriti
  Aggarwal, Owais~Khan Mohammed, Saksham Singhal, Subhojit Som, et~al.
\newblock Image as a foreign language: Beit pretraining for all vision and
  vision-language tasks.
\newblock {\em arXiv preprint arXiv:2208.10442}, 2022.

\bibitem{wang2023images}
Xinlong Wang, Wen Wang, Yue Cao, Chunhua Shen, and Tiejun Huang.
\newblock Images speak in images: A generalist painter for in-context visual
  learning.
\newblock In {\em Proceedings of the IEEE/CVF Conference on Computer Vision and
  Pattern Recognition}, pages 6830--6839, 2023.

\bibitem{wang2023context}
Zhendong Wang, Yifan Jiang, Yadong Lu, Yelong Shen, Pengcheng He, Weizhu Chen,
  Zhangyang Wang, and Mingyuan Zhou.
\newblock In-context learning unlocked for diffusion models.
\newblock {\em arXiv preprint arXiv:2305.01115}, 2023.

\bibitem{wang2021simvlm}
Zirui Wang, Jiahui Yu, Adams~Wei Yu, Zihang Dai, Yulia Tsvetkov, and Yuan Cao.
\newblock Simvlm: Simple visual language model pretraining with weak
  supervision.
\newblock {\em arXiv preprint arXiv:2108.10904}, 2021.

\bibitem{wei2021flan}
Jason Wei, Maarten Bosma, Vincent Zhao, Kelvin Guu, Adams~Wei Yu, Brian Lester,
  Nan Du, Andrew~M Dai, and Quoc~V Le.
\newblock Finetuned language models are zero-shot learners.
\newblock In {\em International Conference on Learning Representations}, 2021.

\bibitem{wu2020phrasecut}
Chenyun Wu, Zhe Lin, Scott Cohen, Trung Bui, and Subhransu Maji.
\newblock Phrasecut: Language-based image segmentation in the wild.
\newblock In {\em Proceedings of the IEEE/CVF Conference on Computer Vision and
  Pattern Recognition}, pages 10216--10225, 2020.

\bibitem{wu2017aic}
Jiahong Wu, He Zheng, Bo Zhao, Yixin Li, Baoming Yan, Rui Liang, Wenjia Wang,
  Shipei Zhou, Guosen Lin, Yanwei Fu, et~al.
\newblock Ai challenger: a large-scale dataset for going deeper in image
  understanding.
\newblock {\em arXiv preprint arXiv:1711.06475}, 2017.

\bibitem{xu2023side}
Mengde Xu, Zheng Zhang, Fangyun Wei, Han Hu, and Xiang Bai.
\newblock Side adapter network for open-vocabulary semantic segmentation.
\newblock In {\em Proceedings of the IEEE/CVF Conference on Computer Vision and
  Pattern Recognition}, pages 2945--2954, 2023.

\bibitem{xu2021}
Mengde Xu, Zheng Zhang, Fangyun Wei, Yutong Lin, Yue Cao, Han Hu, and Xiang
  Bai.
\newblock A simple baseline for open-vocabulary semantic segmentation with
  pre-trained vision-language model.
\newblock In {\em European Conference on Computer Vision}, pages 736--753.
  Springer, 2022.

\bibitem{xu2022vitpose}
Yufei Xu, Jing Zhang, Qiming Zhang, and Dacheng Tao.
\newblock Vitpose: Simple vision transformer baselines for human pose
  estimation, 2022.

\bibitem{yang2023paint}
Binxin Yang, Shuyang Gu, Bo Zhang, Ting Zhang, Xuejin Chen, Xiaoyan Sun, Dong
  Chen, and Fang Wen.
\newblock Paint by example: Exemplar-based image editing with diffusion models.
\newblock In {\em Proceedings of the IEEE/CVF Conference on Computer Vision and
  Pattern Recognition}, pages 18381--18391, 2023.

\bibitem{yang2022unified}
Jianwei Yang, Chunyuan Li, Pengchuan Zhang, Bin Xiao, Ce Liu, Lu Yuan, and
  Jianfeng Gao.
\newblock Unified contrastive learning in image-text-label space.
\newblock In {\em Proceedings of the IEEE/CVF Conference on Computer Vision and
  Pattern Recognition}, pages 19163--19173, 2022.

\bibitem{yang2022lavt}
Zhao Yang, Jiaqi Wang, Yansong Tang, Kai Chen, Hengshuang Zhao, and Philip~HS
  Torr.
\newblock Lavt: Language-aware vision transformer for referring image
  segmentation.
\newblock In {\em CVPR}, 2022.

\bibitem{yildirim2023inst}
Ahmet~Burak Yildirim, Vedat Baday, Erkut Erdem, Aykut Erdem, and Aysegul
  Dundar.
\newblock Inst-inpaint: Instructing to remove objects with diffusion models.
\newblock {\em arXiv preprint arXiv:2304.03246}, 2023.

\bibitem{yu21ap10k}
Hang Yu, Yufei Xu, Jing Zhang, Wei Zhao, Ziyu Guan, and Dacheng Tao.
\newblock Ap-10k: A benchmark for animal pose estimation in the wild.
\newblock In {\em Thirty-fifth Conference on Neural Information Processing
  Systems Datasets and Benchmarks Track (Round 2)}, 2021.

\bibitem{yu2022coca}
Jiahui Yu, Zirui Wang, Vijay Vasudevan, Legg Yeung, Mojtaba Seyedhosseini, and
  Yonghui Wu.
\newblock Coca: Contrastive captioners are image-text foundation models.
\newblock {\em arXiv preprint arXiv:2205.01917}, 2022.

\bibitem{yu2016modeling}
Licheng Yu, Patrick Poirson, Shan Yang, Alexander~C Berg, and Tamara~L Berg.
\newblock Modeling context in referring expressions.
\newblock In {\em Computer Vision--ECCV 2016: 14th European Conference,
  Amsterdam, The Netherlands, October 11-14, 2016, Proceedings, Part II 14},
  pages 69--85. Springer, 2016.

\bibitem{yuan2021florence}
Lu Yuan, Dongdong Chen, Yi-Ling Chen, Noel Codella, Xiyang Dai, Jianfeng Gao,
  Houdong Hu, Xuedong Huang, Boxin Li, Chunyuan Li, et~al.
\newblock Florence: A new foundation model for computer vision.
\newblock {\em arXiv preprint arXiv:2111.11432}, 2021.

\bibitem{zhang2022glipv2}
Haotian Zhang, Pengchuan Zhang, Xiaowei Hu, Yen-Chun Chen, Liunian Li, Xiyang
  Dai, Lijuan Wang, Lu Yuan, Jenq-Neng Hwang, and Jianfeng Gao.
\newblock Glipv2: Unifying localization and vision-language understanding.
\newblock {\em Advances in Neural Information Processing Systems},
  35:36067--36080, 2022.

\bibitem{zhang2023magicbrush}
Kai Zhang, Lingbo Mo, Wenhu Chen, Huan Sun, and Yu Su.
\newblock Magicbrush: A manually annotated dataset for instruction-guided image
  editing.
\newblock {\em arXiv preprint arXiv:2306.10012}, 2023.

\bibitem{zhang2022contrastive}
Yuhao Zhang, Hang Jiang, Yasuhide Miura, Christopher~D Manning, and Curtis~P
  Langlotz.
\newblock Contrastive learning of medical visual representations from paired
  images and text.
\newblock In {\em Machine Learning for Healthcare Conference}, pages 2--25.
  PMLR, 2022.

\bibitem{zhou2017scene}
Bolei Zhou, Hang Zhao, Xavier Puig, Sanja Fidler, Adela Barriuso, and Antonio
  Torralba.
\newblock Scene parsing through ade20k dataset.
\newblock In {\em Proceedings of the IEEE conference on computer vision and
  pattern recognition}, pages 633--641, 2017.

\bibitem{zhu2022uni}
Xizhou Zhu, Jinguo Zhu, Hao Li, Xiaoshi Wu, Hongsheng Li, Xiaohua Wang, and
  Jifeng Dai.
\newblock Uni-perceiver: Pre-training unified architecture for generic
  perception for zero-shot and few-shot tasks.
\newblock In {\em Proceedings of the IEEE/CVF Conference on Computer Vision and
  Pattern Recognition}, pages 16804--16815, 2022.

\end{thebibliography}
}

\end{document}